\documentclass[11pt]{article}
\usepackage{subcaption}

\usepackage{amssymb,amsthm,amsmath,amssymb,wrapfig,dsfont,authblk}
\usepackage{thmtools,thm-restate}
\PassOptionsToPackage{hyphens}{url}
\usepackage{varioref}
\usepackage{verbatim}
\usepackage[usenames,svgnames,xcdraw,table]{xcolor}
\definecolor{DarkBlue}{rgb}{0.1,0.1,0.5}
\definecolor{DarkGreen}{rgb}{0.1,0.5,0.1}
\usepackage[backref=page]{hyperref}
\hypersetup{
     colorlinks   = true,
     linkcolor    = DarkBlue, % color of internal links
     urlcolor     = DarkBlue, % color of external links
	 citecolor    = DarkGreen % color of links to bibliography
}
\renewcommand*{\backref}[1]{}
\renewcommand*{\backrefalt}[4]{%
    \ifcase #1 (Not cited.)%
    \or        (Cited on page~#2)%
    \else      (Cited on pages~#2)%
    \fi}
\usepackage[margin=0.9in]{geometry}
\usepackage{enumerate}
\usepackage{graphicx}
\usepackage{subdepth}
\usepackage{enumitem}
\usepackage{tcolorbox}
\usepackage{algorithm}
\usepackage{algorithmic}
\usepackage{mathtools}
\usepackage{caption}
\usepackage{bbm}
\usepackage{soul}
\usepackage{mathrsfs,amsfonts,dsfont,authblk}
\usepackage{thmtools,thm-restate}
\usepackage[capitalise,noabbrev]{cleveref}
\usepackage[font=footnotesize,labelsep=newline,justification=centering]{caption}
\crefname{property}{Property}{Properties}
\usepackage{nicefrac}
\usepackage{multicol}
\usepackage{enumerate}
\usepackage{libertine}
\usepackage{wrapfig}
\usepackage[backgroundcolor=blue!3!white]{todonotes}
\usepackage[labelfont={normalfont,bf},textfont=it]{caption}
\usepackage{tikz}
\usetikzlibrary{positioning}
\usetikzlibrary{calc}
\newtheorem{lemma}{Lemma}
\newtheorem{claim}{Claim}
\newtheorem{property}{Property}
\newtheorem{definition}{Definition}%[section]

\DeclareMathOperator*{\argmax}{arg\,max}
\newcommand{\ALG}{\textsc{Alg}}

\newcommand{\MAB}{\textsc{MAB}}
\newcommand{\CSUCB}{\textsc{CS-UCB}}

\newcommand{\muvector}{\boldsymbol{\mu}}

\newcommand{\BS}{Bounded smoothness}
\newcommand{\LS}{Lipschitz smoothness}
\newcommand{\R}{R}

\let\oldnl\nl% Store \nl in \oldnl
\newcommand{\nonl}{\renewcommand{\nl}{\let\nl\oldnl}}%

\allowdisplaybreaks

\usepackage{tikz}
\usetikzlibrary{shapes,arrows,decorations.markings}
\usepackage{titlesec}
\titlespacing*{\section}{0pt}{0.5cm}{0.10cm}
\titlespacing{\subsection}{0pt}{2pt}{1pt}
\usepackage{lipsum}

\title{Sleeping Combinatorial Bandits}
\date{ }
\author{Kumar Abhishek, IIIT Hyderabad, India. \hfill  kumar.abhishek@research.iiit.ac.in      
\and Ganesh Ghalme, Technion, Israel. \hfill ganeshg@campus.technion.ac.il \and
 Sujit Gujar, IIIT Hyderabad, India.  \hfill sujit.gujar@iiit.ac.in \and Yadati Narahari, IISc Banglore, India. \hfill narahari@iisc.ac.in \\}

\begin{document}
\maketitle
\begin{abstract}
 In this paper, we study an interesting combination of sleeping and combinatorial stochastic bandits. In the mixed model studied here, at each discrete time instant,  an arbitrary \emph{availability set} is generated from a fixed set of \emph{base} arms. An algorithm can select a subset of arms from the  \emph{availability set} (sleeping bandits) and receive the corresponding reward along with semi-bandit feedback (combinatorial bandits). 
 %The algorithm's performance is evaluated in terms of the regret defined with respect to the best subset of arms from the set of available arms. 
 We adapt the well-known CUCB algorithm in the sleeping combinatorial bandits setting and refer to it as \CSUCB. We prove --- under mild smoothness conditions --- that the \CSUCB\ algorithm achieves an $O(\log (T))$ instance-dependent regret guarantee.  We further prove that (i) when the range of the rewards is bounded,  the regret guarantee of  \CSUCB\ algorithm is  $O(\sqrt{T \log (T)})$  and  (ii)  the instance-independent regret is $O(\sqrt[3]{T^2 \log(T)})$ in a general setting. Our results are quite general and hold under general environments --- such as non-additive reward functions, volatile arm availability, a variable number of base-arms to be pulled --- arising in practical applications. We validate the proven theoretical guarantees through experiments.
 %on simulated data.
 %We also validate our results with extensive simulations with different reward functions.  
\end{abstract}
%%%%%%%%%%%%%%%%%%%%%%%%%%%%%%%%%%%%%%%%%%%
\section{Introduction}
\label{sec:intro}
%%%%%%%%%%%%%%%%%%%%%%%%%%%%%%%%%%%%%%%%%%%

The stochastic multi-armed bandit (MAB) problem is one of the fundamental online learning problems that captures the classic exploration vs. exploitation dilemma. A MAB algorithm, operating in an uncertain environment, is expected to optimally trade-off acquisition of new information with optimal use of information-at-hand to choose an action that maximizes the expected reward or, equivalently, minimizes the expected regret.  In a classical stochastic MAB setup, an  algorithm has to pull (aka select) a single arm (aka choice) at each time instant and receive a reward corresponding to a pulled arm. The reward from each arm is an independent sample from a fixed but unknown stochastic distribution. The goal is to minimize the \emph{expected regret};  the difference between the expected cumulative reward of the best offline algorithm with known distributions and the expected cumulative reward of the algorithm.

% \citet{auer2002} propose the famous  UCB1 algorithm and show that it achieves asymptotically optimal (up to instance dependent constants) regret; i.e., attains  $O(\log T)$ regret, where $T$ is the number of time steps.  \citet{agrawal12} shows that the regret of the Thompson sampling (TS) algorithm \cite{thompson1933likelihood} also attains optimal regret guarantee with smaller instance dependent constants. Though theoretically, both algorithms have comparable regret bounds, UCB1 is easier to analyze and has been widely used. 

In this paper, we study a combination  of two well studied extensions of classical stochastic MABs, namely \emph{sleeping bandits} \cite{KLE10} and \emph{combinatorial bandits} \cite{gai12}. In the sleeping bandits setting, only a subset of base arms is available at each time instant. This variant, sometimes also known as volatile bandits  \cite{bnaya2013volatile} or mortal bandits \cite{deepNIPS08}, models many real-world scenarios such as crowdsourcing \cite{li2019combinatorial}, online advertising \cite{deepNIPS08}, and network routing \cite{KLE10,bnaya2013volatile} where an algorithm is restricted to  select from only the available set of choices. 

Another well studied generalization of the classical MAB setting is the combinatorial MAB (CMAB) problem    \cite{audi09,gai10,bianchi12cmab,combes_NIPS2015,WANG18}.  Similar to sleeping bandits, this variant too provides an abstraction to many real-world decision problems. For instance, in an online advertising setup, the platform selects multiple ads to display at any point in time \cite{gai10}; in crowdsourcing, the requester chooses multiple crowd workers at the same time \cite{ul2016efficient} and in network routing algorithm has to choose a path instead of a single edge \cite{talebi2017stochastic,kveton2015combinatorial}. Studying the two settings together presents interesting and non-trivial technical challenges. %\ka{Should we highlight the applications of general reward functions?}

We consider the semi-bandit feedback model and a general reward function (under mild smoothness constraints). In the semi-bandit feedback model, an algorithm observes the reward realizations corresponding to each of the selected arms along with the overall reward for pulling the subset of arms. The smoothness properties on the reward functions studied in this paper are similar to those in \cite{CHE13}. It is worth mentioning here that in the sleeping MAB setting, the conventional definition of regret is not appropriate as the best arm (or the best subset of arms in the combinatorial sleeping MAB case) may not be available at all time instants. Hence, we evaluate the performance of an algorithm in terms of its  \emph{sleeping regret}  \cite{KLE10},   defined as the difference between the expected reward obtained from best \emph{available} arm and the arm pulled by the algorithm.

%\textcolor{red}{Applications 1) Stochatic online routing with inactive nodes 2) online advetisersement and recommendation systems with unvailable advertisers 3)  \cite{talebi2017stochastic} }

The paper is organized as follows: 
In Section \ref{sec:model}, we formally introduce the sleeping combinatorial bandits problem and define Lipschitz smoothness and Bounded smoothness assumptions. The required notational setup is introduced  and \CSUCB\ algorithm is given in Section \ref{sec:setting}. In Section \ref{sec:theory results}, we  provide regret analysis of  \CSUCB\ under \LS\ and \BS\ assumptions. In Section \ref{sec:simulation}, we provide an in-depth verify of the theoretical results on simulated data with few reward functions. The related literature is discussed in  Section \ref{sec:reltd_wrk} and in Section \ref{sec:conclusion},  we conclude our paper with a brief discussion on the results and future directions. 

%%%%%%%%%%%%%%%%%%%%%%%%%%%%%%%%%%%%%%%%%%%%%%%%%%%%%%%%%%%%%%%%
\section{Model and Assumptions}
\label{sec:model}

 In a classical stochastic multi-armed bandits (\MAB) problem, at each discrete time step $t$, an algorithm  pulls a single arm $i_t \in [k]$  and observes a random reward $X_{i_t, t}$.  The random variables $(X_{i,t})_t$ are identical and independently distributed according to a distribution $\mathcal{D}_i(\mu_i)$. Here, $\mu_i$ is the mean of distribution $\mathcal{D}_i$. Note that the reward  corresponding to  arms $j \neq i_t$ is not observed. %We consider that $\mathcal{D}_i$ is supported over $[0,1]$.
%\footnote{We have assumed this for ease of exposition. However, all the results can be extended to any distribution over a finite support. \textcolor{red}{cite the paper}}
The reward distributions $(\mathcal{D}_i)_{i \in [k]}$ are unknown to the algorithm. Throughout this paper we consider that the reward distributions have a bounded support.  The algorithm's objective is to minimize expected regret defined as, $\mathcal{R}_{\ALG}(T)  = \mathbbm{E} [ \sum_{t=1 }^{T} (X_{i^{\star},t} -  X_{i_t, t})]$. Here, $i^{\star} = \arg\max_{i} \mu_i$ denotes the best arm.  
%Note that the time horizon $T$ is a-priori unknown to the algorithm \footnote{For horizon aware algorithms refer to \cite{bubeck12}.}. 

In this paper we consider a sleeping combinatorial bandits problem with  $[k]:= \{1,2, \cdots, k\}$ denoting  the set of \emph{base} arms and $\muvector \in [0,1]^{k}$,  the vector of unknown mean qualities of the base arms. Similar to the classical stochastic MAB problem, each base arm $i$ corresponds to an unknown  distribution $\mathcal{D}_{i} $ with mean $\mu_i \in [0,1]$ over its quality. At each time instant $t$, a subset $A_t\subseteq [k]$ of the base arms become available. Throughout the paper, we consider that $A_t$ is an arbitrary non-empty subset. A  decision maker (i.e. an algorithm) can pull any non-empty subset $S_t \subseteq A_t$ of arms and receive a  reward $\R_{t} := \R(S_t,\muvector)$.  The reward depends upon the  selected subset $S_t$ and the mean qualities of the arms, $\muvector$. We define, $\R_S := R(S,\muvector)$ whenever the quality vector is clear from the context. Furthermore, the reward depends only on the qualities of  pulled arms $S_t$ \footnote{That is for any set $S \subseteq [k]$, $\R_S(\muvector) = \R_{S}(\muvector^{'})$ if $\muvector_{i} = \muvector_{i}^{'}$ for all $i \in S$. }.   
 We remark here that the classical stochastic bandits setting is a special case of our setting with $A_t = [k], |S_t|=1$ and $R_t = X_{S_t,t}$ for all $t$.

For a given  reward function  $\R$ %: 2^{[k]} \rightarrow \mathbbm{R}_{+}$, 
the problem reduces to finding a reward maximizing subset of arms.  This problem, even when the qualities of the base arms are known,   is known to be NP-hard in general \cite{wolsey1999integer}. However,  many important settings, such as  submodular reward functions, admit  a polynomial time approximation schemes that  provides  a decent approximation guarantee.  
To demarcate the computational problem of finding an optimal set of arms from effectively learning the quality  distributions (and hence the learning an optimal set of arms to be pulled)  we assume the existence of an $(\gamma, \beta)$-approximation oracle (denoted by   \textsc{$(\gamma, \beta)$-Oracle} ),  which,  given an availability set $A$ and a  quality vector $\muvector$,  outputs  a set $S$ such that $\R_{S}(\muvector)\geq \gamma \cdot \R_{S^{'}}(\muvector ) $,  for all $S^{'} \in 2^{A}$  with the probability of at least $\beta$, with $  \gamma, \beta \in (0,1]$.
\iffalse
as below. 
\iffalse
\begin{flushleft}
\fbox{
\begin{minipage}[5cm][2.1cm][t]{
0.9 \columnwidth} 

\textsc{($\gamma, \beta$)-Oracle} \newline  ---------------------- \newline 
\textbf{Input} $\muvector , A; \gamma, \beta \in (0,1]$  \newline 
\textbf{Output} $S \subseteq A$ such that $\mathbbm{P}\big \{ \R_{S}(\muvector) \geq \gamma \cdot \R_{S^{\star}}(\muvector)\big \} \geq \beta$. Here, $S^{\star} \in \arg\max\limits_{S' \subseteq A} \R_{S^{'}}(\muvector)$  
\end{minipage} }
\end{flushleft} 
\fi
\begin{align*}
\textsc{\textsc{($\gamma, \beta$)-Oracle}}~~~& \textbf{Input: } \muvector , A; \gamma, \beta \in (0,1]  \\
 &\textbf{Output: } S 
 \text{ such that, } \\ 
 & \mathbbm{P}\big \{ \R_{S}(\muvector) \geq \gamma \cdot \R_{S^{\star}}(\muvector)\big \} \geq \beta.
\end{align*}
\fi 
The  computation oracle separates the learning task from the offline computation task and is extensively used in the literature \cite{gai12,CHE13,chenNIPS16}.

For the semi-bandit feedback to work effectively,  we assume some smoothness properties on the reward function.  These smoothness properties ensure that when the learning parameters are estimated with a certain precision, one can approximate the true reward with high accuracy. Formally,  the reward function $\R_{S}(\muvector)$, as a function of stochastic parameter $\muvector$, satisfies the following properties. 

 \begin{property}
 \label{prop:mono}
 \textbf{Monotonicity}: Let $\boldsymbol\mu , \boldsymbol \mu^{'} \in [0,1]^k$ be
     two vectors such that $ \muvector_i^{'} \geq \muvector_i $ for all $i \in [k]$ then, for any $S \subseteq [k]$,    $\R_{S}(\muvector^{'}) \geq  \R_{S}(\muvector)$.
 \end{property}
 The monotonicity property implies that the reward from any subset increase if the mean qualities of an base arms increase. 
 \begin{property}
\label{prop:lipschitz}
\textbf{Lipschitz Continuity}: There exists real valued constant $C \geq  1$ such that for all $ S \subseteq [k]$, we have $|\R_{S}(\muvector) - \R_{S}(\muvector^{'}) |\leq C \max_{i \in S}|\muvector_i - \muvector_i^{'}|$.
\end{property}

 \begin{property}
 \label{prop:smooth}
 \textbf{Bounded Smoothness}: There exists a strictly increasing function $f$ such that for any $S \subseteq [k]$,  $|\R_{S}(\muvector) - \R_{S}(\muvector^{'}) |\leq f(\Lambda)$ whenever $\max_{i \in S }|\muvector_i - \muvector_{i}^{'} | \leq \Lambda$.
 \end{property}

In our first setting we study  the reward function  $\R(\cdot)$ satisfying monotonicity (Property \ref{prop:mono}) and the Lipschitz continuity property (Property \ref{prop:lipschitz}) whereas in the setting setting we consider Property \ref{prop:mono} and Property \ref{prop:smooth}. With a slight abuse of terminology, we call the first setup as \LS\ and the second setup (i.e. monotonicity and bounded smoothness) as the \BS.

The reward assumptions and regret notion considered in the paper encompass many specialized settings studied in literature as a special case. For instance,  additive rewards with a fixed number of arms to pull \cite{kveton2015}, submodular rewards with volatile bandits \cite{chen2018contextual},  average reward, and so on. 
%The problem of  for the stochastic combinatorial sleeping MAB is an open problem to the best of our knowledge. We address this problem partially in this paper. 
%We adopt the CUCB algorithm \cite{CHE13} in the sleeping CMAB setting and call it a CS-UCB.
However, we remark here that 
the technical treatment of this problem requires newer proof techniques as the existing proof techniques from combinatorial bandits setup do not generalize trivially to the sleeping combinatorial bandits setting.
\subsection*{Main Results of the Paper}
\begin{itemize}[noitemsep,leftmargin=*]
%    \item We adopt the CUCB algorithm \cite{CHE13} to the combinatorial sleeping MAB setting as \CSUCB. When all the arms are available at all time instances, \CSUCB\ is the same as CUCB.
    \item In the Lipschitz smoothness setting, we show  that \CSUCB\  achieves   $O(\log(T)/\Delta_{\min})$ instance-dependent  regret guarantee (See Theorem \ref{thm:Regret_CSUCB_lipschitz}).  Here, $\Delta_{\min}$ is the difference between the reward from an optimal super-arm and a sub-optimal super-arm with maximum reward. 
    \item Note that  for smaller values of $\Delta_{\min}$, the regret guarantee of Theorem \ref{thm:Regret_CSUCB_lipschitz} regret guarantee is vacuous. In Theorem \ref{thm:Regret_CSUCB_independent_BoundedMax}, we show that CS-UCB attains an instance-dependent regret guarantee of  $O(\sqrt{\sigma k T \log(T)})$. Here, $\sigma = \Delta_{\max}/\Delta_{\min}$. Note that, in contrast with Theorem \ref{thm:Regret_CSUCB_lipschitz} this result depends only on the range of rewards of super-arms. In particular, if the best and worst super-arms do not have a large reward ratio, the result in Theorem \ref{thm:Regret_CSUCB_independent_BoundedMax} is tight. We refer to this setting as \emph{weak instance-dependent}.
    
    \item Next, in Theorem \ref{thm:Regret_CSUCB_independent}, we obtain a $O(\sqrt[3]{kT^2\log(T)})$ instance-independent regret guarantee  without any dependence on $\sigma$ in a Lipschitz smoothness setting.
    
    \item Finally, in a Bounded smoothness setting, in Theorem \ref{thm:Regret_CSUCB1_bounded}, we show that \CSUCB\ attains $O(\log(T))$  regret guarantee. Though a similar result exists for the non-sleeping case \cite{CHE13}; the regret analysis does not trivially generalize to the combinatorial sleeping MAB setting.
\end{itemize}

%%%%%%%%%%%%%%%%%%%%%%%%%%%%%%%%%%%%%%%%%%%
\section{The Setting}
\label{sec:setting}
%%%%%%%%%%%%%%%%%%%%%%%%%%%%%%%%%%%%%%%%%%%

In this paper,  we consider that only a subset  $A_t \subseteq [k]$  of arms is available at time $t$.  
Note that, $A_t$ is revealed only at time $t$. Further, let $S_t \subseteq A_t$ be the set of arms pulled by the algorithm at time $t$. The set $S_t$ is also called as a \emph{super-arm}. 
To evaluate the performance of an algorithm with limited availability of arms,  we extend the notion of regret considered for classical CMAB problem appropriately and call it a \emph{sleeping regret} given by $
    \mathcal{R}_{\ALG}(T) := \max_{(A_t)_{t=1}^{T}} \mathbbm{E}_{\ALG} \big [  \sum_{t=1}^T ( \R_{S_{t}^{\star}}  - \R_{S_t} ) \big ] . 
$
Here, $S_t^{\star} \in \arg \max_{S \subseteq A_t} \R_{S} $. Note that when $A_t = [k]$ for all $t$, we recover the setting of \cite{CHE13}. 
\iffalse
The objective of the algorithm is to minimize the expected cumulative sleeping regret.
Notice that the algorithm pulls the super-arm $S_t$ based on the UCB estimates $\overline{\boldsymbol{\mu}}_t $. Here $S_{\muvector_t}^{\star}(A_t)$ denotes the optimal super-arm within available set $A_t$, i.e., $S_{\muvector_t}^{\star}(A_t) \in \arg\max_{S \subseteq A_t }\R_{\boldsymbol{\mu_t}}(S) $. 
\fi 
 Next, we define the regret in the presence of $(\gamma, \beta)$-oracle. Let, $B_t$ be the event that an oracle returns an  $\gamma$-approximate solution at time $t$ i.e. $B_t =  \{\R_{S_t} \geq \gamma \cdot \R_{S_{t}^{\star}} \}$.  Note that $\mathbbm{P}(B_t) \geq \beta$. The expected sleeping regret of \ALG\ with oracle access is  given by, 
\begin{align}
\mathcal{R}_{\ALG}(T) &= \max_{(A_t)_{t=1}^{T}} \mathbbm{E}_{\ALG} \Big [  \sum_{t=1}^T ( \gamma \cdot \beta \cdot  \R_{S_{t}^{\star}} - \R_{S_t} ) \Big ] .
\label{eqn:sleepingRegret}
\end{align}
\subsection*{Notational Setup}
We begin with the additional notation  required to prove the results.  
For each base arm  $i \in [k]$, let  $N_{i,t}$ denotes the number of times arm $i$ is pulled till time $t$ and $\hat{\mu}_{i,t}$ be the  average reward obtained from arm $i$ till (and excluding) time $t$.  Let
\begin{equation}
\label{eqn:UCBEstimate}
\overline{\mu}_{i,t} := \hat{\mu}_{i,t} + \sqrt{3\log(t)/2N_{i,t}}.
\end{equation}
Following a standard terminology, we call $\overline{\mu}_{i,t}$ as the UCB estimate of arm $i$ at time $t$.  
Furthermore, let $\Delta_{S} := \gamma \cdot \textsc{opt}_{A} - \R_{S}$ be the regret incurred by pulling super-arm $S$. Here,   $ \textsc{opt}_{A}  := \R_{S^{\star}}  = \max_{S \subseteq A} \R_{S} $ denotes the optimal reward when the set of available arms is $A$. A super-arm $S \subseteq A $ is \emph{bad} (sub-optimal), if $\Delta_{S} > 0$. 
For a given $A \subseteq [k]$, we define the set of bad super-arms as $S_B(A) = \{S \subseteq A | \Delta_{S} > 0 \}$.  Further, for a given $A \subseteq [k]$,  define 
\begin{align*}
        \Delta_{\min}(A)  &= \gamma \cdot \textsc{opt}_{A} - \max_{S  \in S_{B}(A)} \R_{S} \text{ \ \ \ \ and, } \\ 
        \Delta_{\max}(A)  &= \gamma \cdot \textsc{opt}_{A} - \min_{S  \in S_{B}(A)} \R_{S}.
\end{align*}
Note that, for any availability set $A$, we have $  \Delta_{\max}(A) \geq \Delta_{\min}(A) > 0$. The strict inequality follows from the definition of $S_{B}(A)$.
Next, define   $\Delta_{\max} = \max_{A \subseteq [k]} \Delta_{\max}(A) $ and $\Delta_{\min} = \min_{A \subseteq [k]} \Delta_{\min}(A) $. 
%Without loss of generality, we assume that the reward function $r(.)$ is bounded i.e. there exists $\alpha > 0 $ such that $r_{\muvector}(S) \leq \alpha $ for all $\muvector \in [0,1]^{k}$ and for all $S \in [k]$. Notice that $\Delta_{\max} \leq \alpha$.  
\iffalse
\begin{figure*}[ht!]
    \centering
    \includegraphics[width=1.5\columnwidth,height=6cm]{Time_div.png}
    \caption{Division of total time instants $T$ in $T_e$ and $T_u$ sets}
    \label{fig:csucb_timediv}
\end{figure*}
\fi
%Let $(A_t, S_t)_{t=1}^{T}$ denote a  sequence of available arms  $\{A_t\}$  and   the set of arms pulled by  \textsc{CS-UCB} $\{S_t\}$. 
Arm $i$ is called \emph{saturated} if it is pulled for sufficiently many number of time steps, i.e., $N_{i,t} \geq \ell_t$, where $\ell_t$ works as threshold exploration. Note that a saturated arm at any time instant may become unsaturated in future. Further, we call a set $A_t$ \emph{explored} if all the arms in $A_t$ are saturated, i.e., $N_{i,t} \geq \ell_t $ for all $i\in A_t$. 
First observe that if $ A_t $ is either empty or a singleton set then \CSUCB\ incurs a zero regret. Hence, without loss of generality we assume that $|A_t| \geq 2$ for all $t \leq T$.  We first make following useful observation.

\begin{restatable}{observation}{ObservationOne}
\label{obs:One} 
Let $\ell_t \in \mathbbm{R}_{+}$ be a positive number,  $S \subseteq [k]$ be any non-empty set of arms and such that $N_{i,t} \geq \ell_t $ for all $i \in S$ and  $\varepsilon_{t} :=  \sqrt{\frac{3\log (t)}{2 \ell_t}}  $,  then     $\mathbbm{P}\{ \max_{i \in S} | \overline{\mu}_{i,t} - \mu_i | < 2 \varepsilon_{t} \} \geq 1- 2|S|/t^3.$
\end{restatable}
The proof of Observation \ref{obs:One} follows from Hoeffding's inequality and is presented in the supplementary material for completeness.

\subsection*{ \CSUCB\:  }
\label{sec:theory}
%%%%%%%%%%%%%%%%%%%%%%%%%%%%%%%%%%%%%%%%%%%%%%%%%%%%%%%
%We now present  \CSUCB\ algorithm. 
%Next, we  prove the regret upper bounds under \LS\ and \BS\ settings. 
Note that the proposed \CSUCB\  algorithm is the same as CUCB \cite{CHE13} except that at each time, only a subset of the arms is available, and the regret notion considered is sleeping regret instead of conventional regret.     Similar to CUCB, we assume that the algorithm has access to a  $(\gamma, \beta)$-approximation oracle. 

At each time $t$, \CSUCB\ receives the set of available arms $A_t$. If there is a base arm in $A_t$ which is not pulled previously, an algorithm pulls all the available arms.  For each time instances where all available arms are pulled atleast once,  \CSUCB\  obtains  $S_t = \textsc{Oracle}(\overline{\muvector}_{t}, A_t )$. Here, $\overline{\muvector}_t$ represent the vector of UCB estimates given by Equation \ref{eqn:UCBEstimate}. The algorithm then pulls a super-arm $S_t$ and obtain rewards $R_{S_t}(\muvector)$ and an individual base arm rewards (semi-bandit feedback) $X_{i,t}$ for each $i \in S_t$. Finally, \CSUCB\ update parameters \begin{itemize} \item $N_{i,t+1} = N_{i,t} + \mathbbm{1}(i \in S_t)$ \item   $\overline{\mu}_{i,t+1} = \frac{N_{i,t} \cdot \hat{\mu}_{i,t} + \mathbbm{1}(i \in S_t) \cdot X_{i,t}}{ N_{i,t} + \mathbbm{1}(i \in S_t)} + \sqrt{\frac{3 \log(t)}{N_{i,t} + \mathbbm{1}(i \in S_t)}} $.     
\end{itemize}

Note that the regret (Equation \ref{eqn:sleepingRegret})  depends on the rewards from the base arm $X_{i,t}$ only through $R(.)$.  Further, observe that  when $A_t = [k]$  for all $t$, the sleeping regret is same as conventional regret guarantee and \CSUCB\ is same as CUCB; hence  the regret guarantees of \cite{CHE13} will apply. 

%$\mathbbm{P} \{ R_{S_t}(\muvector; \overline{\muvector})  = R_{S_t}(\muvector; \muvector)  \}$

%%%%%%%%%%%%%%%%%%%%%%%%
\section{Regret Analysis of \CSUCB\  }
\label{sec:theory results}
In our first result, we show that under the Lipschitz smoothness setting, \CSUCB\ incurs a logarithmic instance-dependent regret. However, note that the regret depends inversely on the $\Delta_{\min}$ value. That is, for arbitrarily smaller values of $\Delta_{\min}$, the regret bound is vacuous. In Theorem \ref{thm:Regret_CSUCB_independent_BoundedMax}, we prove that the weak instance-dependent regret of the proposed algorithm is $O(\sqrt{\sigma kT\log(T)})$. Here, $\sigma =\Delta_{\max} / \Delta_{\min}$; i.e., this result depends only on the ratio of the maximum and minimum achievable rewards. Finally, in Theorem \ref{thm:Regret_CSUCB_independent}, we show that the instance-independent regret of the proposed algorithm is $O(\sqrt[3]{k T^2 \log(T)})$ in general. We begin with the following observation. 

\begin{restatable}{observation}{ObservationTwo}
\label{obs:second} For all time instances $t$  such that $N_{i,t} > 0$ and the reward function satisfies monotonicity  and  Lipschitz continuity (Properties  \ref{prop:mono} and \ref{prop:lipschitz}),
$\Delta_{S_t} \leq C \big( 1 + \sqrt{3 \log(T)/2}\big).$
\end{restatable}
\noindent We are ready to present out first result. 
%\subsection{Instance-dependent sleeping regret} 
\begin{restatable}{theorem}{RegretLipschitzDependent}
\label{thm:Regret_CSUCB_lipschitz}
The expected sleeping regret incurred by \CSUCB\ when the reward function satisfies   Lipschitz condition (Properties \ref{prop:mono} and  \ref{prop:lipschitz})  is given   by 
\begin{equation*}
\tiny 
\mathcal{R}_{\CSUCB} (T) \leq 2 \beta k C  \left[ \zeta(3) ( 1 + \sqrt{\frac{3 \log(T)}{2}})  +  3 \frac{\sigma C \log(T)}{ \Delta_{\min}}  \right]
\end{equation*}
Here, $\zeta$ is the Reimann zeta function and $\sigma = \Delta_{\max}/\Delta_{\min}$.
\end{restatable}
\begin{proof}[Proof Outline:]
 Set $\ell_t := 6C^2\log(t)/\Delta_{\min}^2$ and $\varepsilon_t := \sqrt{3 \log(t)/2 \ell_t} $ and divide the time instants into sets  $T_e$ and $T_u$ as described as follows. Let  $T_e$ be the set of time instances $t$ such that  $A_t$ is  explored, i.e., $T_e = \{ t \leq T : N_{i,t} \geq \ell_t ,  \forall i  \in A_t \}$ and $T_u = [T] \setminus T_e$.  Further let, for $t \in T_u$, $A_{e,t}$ be the  set of saturated arms that are available at time $t$, i.e., $A_{e,t} := \{i | N_{i,t} \geq \ell_t\}$ and  $A_{u,t} := A_{t} \setminus A_{e,t}$. We have
 \begin{align*}
 T_u &= \{t: \exists j \in A_{u,t}  \} \\ 
  &=   \underbrace{ \{t: \exists j  \in A_{u,t}\cap  S_t  \}}_{D}  \cup  \underbrace{\{t: \forall j \in A_{u,t}  , j\notin S_t  \} }_{E}.
 \end{align*}
 We bound the sleeping regret incurred in  disjoint sets $T_{e}$, $D$ and $E$ separately. Recall that $B_t$ is an event that the oracle returns $\gamma$-approximate solution i.e. $\R_{S_t}(\overline{\muvector}_t) \geq \gamma \cdot \R_{S^{\star}}\{ \overline{\muvector}_t \}$.  We begin with  following  supporting lemmas.
\begin{restatable}[]{lemma}{ExploredUpperBoundLS}
\label{lem: first lemma of theorem 1}
For all $t \in T_e$ we have
$
    \mathbbm{P}\{ S_t \in S_{B}(A_t) | B_t \} \leq 2 |A_t| t^{-3}.
$
\label{lem:lemmaOne}
\end{restatable}
\begin{restatable}{lemma}{LemmaCardinalityOfD}
\label{lem:cardinality of D }
$|D| \leq k \ell_{T}$.
\end{restatable}
\begin{restatable}{lemma}{TheoremOneLemmaThree}
 \label{lem:third lemma of theorem 1}For all $t \in E$ we have, 
 $\mathbbm{P}\{S_t \in S_B(A_t)|B_t \} \leq 2|S_t|/ t^3 . $
\end{restatable}
Lemma \ref{lem: first lemma of theorem 1} establishes that when all the base arms in the availability set are sufficiently explored then the set $S_t$ returned by the oracle is an optimal set with high probability. This result follows from the fact that, as all the available arms are sufficiently pulled in the past,  $\overline{\muvector}$ is sufficiently close to $\muvector$. Lemma \ref{lem:cardinality of D } follows directly from the fact that each base arm remains unsaturated till atmost  $\ell_{T}$ pulls.

Finally, in Lemma \ref{lem:third lemma of theorem 1} we handle the case that the availability set contains both saturated and unsaturated base arms. Note that the previous two lemmas also hold for CMAB settings. However, in contrast with CMAB, in our setting, the availability sequence may be such that at each time instant only a few explored arms are available and this may lead to high regret. Lemma \ref{lem:third lemma of theorem 1} dismisses this hypothesis. First, note that only those base arms that are  \emph{available}  but  \emph{not-pulled} are responsible for the regret.  Furthermore, if an arm is available and it is not pulled for many time instances, its UCB estimate increases and hence increasing its chances of getting pulled in the future due to the monotonicity assumption. This means that an optimal subset of the availability set will be pulled after some time with high probability.  The detailed proof of Lemmas \ref{lem: first lemma of theorem 1}, \ref{lem:cardinality of D }
 and \ref{lem:third lemma of theorem 1} are given in supplementary material.
 
 \textbf{Putting everything together:} For a given arbitrary availability sequence $(A_t)_{t=1}^{T}$, the regret of \CSUCB\ is given as 
\begin{align*}
\allowdisplaybreaks
\tiny 
 &\mathcal{R}_{\CSUCB}(T) = \mathbbm{E} \big[\sum_{t\in [T]} \gamma \cdot \beta \cdot \R_{S_t^{\star}} - \R_{S_t}  \big ]\\
 \leq & \mathbbm{E} \big [ \sum_{t\in [T]} \gamma \cdot  \R_{S_t^{\star}} - \R_{S_t}  \mid B_t \big] \cdot \beta  \\ 
 \leq &   \big [ \sum_{t\in T_e \cup E} \mathbbm{P} \{ S_t \in S_{B}(A_t) | B_t \} \cdot  \Delta_{S_t} \\  &+  
        \sum_{t\in D} \mathbbm{P} \{ S_t \in S_{B}(A_t)| B_t \} \cdot \Delta_{\max}  \big ] \cdot \beta \\  
     \leq &  \big [  \sum_{t \in T_e \cup E}  2\frac{|A_t|}{ t^3} \Delta_{S_t} + k \ell_T \Delta_{\max} \big ] \cdot \beta    \tag{From Lemmas \ref{lem: first lemma of theorem 1}, \ref{lem:cardinality of D } and  \ref{lem:third lemma of theorem 1}}\\
      \leq &  \big[ 2k C  \big(1 + \sqrt{\frac{3 \log(T)}{2}}\big) \sum_{t = 1}^{\infty} 1/t^3 + \frac{6k C^2\log(T)}{\Delta_{\min}^2} \cdot \Delta_{\max} \big]  \cdot \beta \tag{from Observation \ref{obs:second}} \\
      \leq & \Big[ 2 k C  \zeta(3) \Big( 1 + \sqrt{3\log(T)/2}\Big) + \frac{6C^2 k \sigma \log(T)}{\Delta_{\min}} \Big]. \beta  
  \end{align*}
 Since the last equation holds for any arbitrary sequence $(A_t)_{t=1}^T$, it also holds for an adversarially chosen availability sequence. This completes the proof of the theorem.
%  The sleeping regret of \CSUCB\ can be upper bounded as 
 % \begin{align*}
%      \mathcal\mathcal{R}_{\CSUCB}(T) \leq & \sum_{t\in T_e \cup E} \mathbbm{P}(S_t \in S_{B}(A_t)) \Delta_{\max} \\ 
%      & + \sum_{t\in D} \mathbbm{P}(S_t \in S_{B}(A_t)) \Delta_{\max} \\
  %    \leq & \sum_{t=1}^T %\frac{\Delta_{\max}}{t^2}|S_t| + \Delta_{\max}|D| \tag{from Lemma \ref{lem:lemmaOne} and Lemma \ref{lem:lemmaCase2}}\\
%      \leq & \left(\frac{\pi^2}{6} +1\right)k \Delta_{\max} + k\ell \Delta_{\max} \tag{ from \ref{case_1}}\\
 %     = & \Big[ \Big(\frac{\pi^2}{6} +1\Big) + \frac{6\log(T)}{(f^{-1}(\Delta_{\min}))^2}  \Big]k\Delta_{\max} \\
 %      \leq & \Big[ \Big(\frac{\pi^2}{6} +1\Big) + \frac{6\log(T)}{(f^{-1}(\Delta_{\min}))^2}  \Big]k\alpha
 % \end{align*}
\end{proof}  
%We note here that the sleeping regret guarantee provided by Theorem \ref{thm:Regret_CSUCB_lipschitz} matches with the regret guarantee given by \citet{CHE13} for the non-sleeping case.

\iffalse
  \begin{restatable}[]{corollary}{nonSleeping}
  \label{cor:cucb_recover}
  If all the arms are available for each round, i.e., $A_t = [k]$ for all $t \leq T$ the upper bound on the regret of \CSUCB\ is given by 
$$   \mathcal{R}_{\CSUCB}(T) \leq \left[  \frac{6\log(T)}{(f^{-1}(\Delta_{\min}))^2} + \frac{\pi^2}{6} + 1\right] k \cdot \alpha$$  \end{restatable}
  Proof of the theorem is given in the Appendix \ref{appndx:missing}. Note that the upper bound  is same in Theorem \ref{thm:Regret_CSUCB1} and Corollary \ref{cor:cucb_recover}, however in Theorem \ref{thm:Regret_CSUCB1} we consider sleeping regret whereas in Corollary \ref{cor:cucb_recover} we consider the conventional notion of regret. The matching bound in both settings is due to the fact that the unavailability of any arm in some time instants can only delay its chance to get pulled and hence delays its learning. However, this delay does not cause additional regret as the regret is computed only with respect to all available arms. We also note here that the analysis of \cite{CHE13} does not handle the unavailability of the arms.
  \fi
Notice that the regret guarantee in Theorem \ref{thm:Regret_CSUCB_lipschitz}  depends on the value of $\Delta_{\min}$. If this value is sufficiently low the regret guarantee is vacuous. In the next result, we show a weak instance-dependent regret guarantee where the regret is given in terms of the ratio  $\Delta_{\max}/\Delta_{\min}$. 
\begin{restatable}{theorem}{InstIndOne}
\label{thm:Regret_CSUCB_independent_BoundedMax}
The weak instance-dependent sleeping regret of \CSUCB\ when the reward function satisfies  Lipschitz condition (properties \ref{prop:mono} and  \ref{prop:lipschitz})  is given by 
$$
      \mathcal{R}_{\CSUCB}(T) \leq 4 C \sqrt{6k\sigma T\log(T)} + 2 k C \zeta(3) .
   %\leq 2 C \cdot \sqrt{6\sigma k T\log(T) } \big[ 1 + \zeta(3) k \sigma \big]
   %2 C \cdot \sqrt{6k \sigma T\log T}    + \left(\zeta(3) \right)Ck
$$
  Here, $\zeta(.)$ is a Reimann zeta function and $ \sigma = \Delta_{\max} / \Delta_{\min}$.
\end{restatable}
It is easy to see that for large values of  $\Delta_{\min}$ one can use the result of Theorem \ref{thm:Regret_CSUCB_lipschitz} to obtain the desired bound  of Theorem \ref{thm:Regret_CSUCB_independent_BoundedMax}. However, when $\Delta_{\min}$ is small i.e. $\Delta_{\min} < C \sqrt{6k\sigma \log(T)/T}$,   the upper bound on regret is obtained by parametrized analysis with selecting parameter $\eta \in (\Delta_{\min}, \Delta_{\max}]$ appropriately to minimize the regret. The detailed proof of Theorem \ref{thm:Regret_CSUCB_independent_BoundedMax} is given in supplementary material.
%Appendix \ref{appndx:missing}. 
Observe that the regret dependence of Theorem \ref{thm:Regret_CSUCB_independent_BoundedMax} on time horizon increase from $O(\log(T))$ to $O(\sqrt{T\log(T)})$ when we consider the weak instance-dependent regret guarantee. In the next result, we further relax the dependence on instance parameters ($\sigma$) to obtain a \emph{strong}   instance-independent regret guarantee of $O(\sqrt[3]{T^2\log(T)})$.
% In the next theorem, we relax this restriction on the maximum and minimum possible rewards.   
\begin{restatable}[]{theorem}{RegLipIndGen} 
\label{thm:Regret_CSUCB_independent}
The instance-independent sleeping regret of \CSUCB\ when the reward function satisfies  Lipschitz condition (Properties \ref{prop:mono} and  \ref{prop:lipschitz})  is given by 
$$ \mathcal{R}_{\CSUCB}(T) 
   \leq C(1+ \lambda ) \cdot \sqrt[3]{6k T^2\log (T)} + 2k \lambda C\zeta(3)$$
  where $\lambda = ( 1 + \sqrt{3 \log(T)/2})$.  
\end{restatable}
%A detailed proof is provided in the supplementary material.
%Appendix \ref{appndx:missing}. 
%The proof follows the proof of Theorem  \ref{thm:Regret_CSUCB_independent_BoundedMax}. %\textcolor{red}{In my opinion its a best idea to keep the proof of theorem 3 and move proof of Theorem 2 to appendix. @Abhishek: can you take care of this? } 
First, using  Theorem \ref{thm:Regret_CSUCB_lipschitz} we establish that the said  instance-independent  upper bound holds in this setting if $\Delta_{\min} \geq (\frac{T}{\log(T)})^{-1/3}$. Then,  similar to Theorem \ref{thm:Regret_CSUCB_independent_BoundedMax}, we split the regret at any time $t$ into two parts, where the per time  regret is at most $\eta$ and larger than $\eta$ with $\eta \geq (\frac{T}{\log(T)})^{-1/3}$.
 As stated previously, this result provides the instance-independent regret guarantee without any additional restrictions on minimum and maximum rewards. 

 \begin{restatable}[]{theorem}{RegretBS}
\label{thm:Regret_CSUCB1_bounded}
The expected sleeping regret incurred by \CSUCB\ when the reward function satisfies bounded smoothness  condition (Properties \ref{prop:mono}  and  \ref{prop:smooth}),   is upper bounded by 
$$\mathcal{R}_{\CSUCB} (T) \leq \Big[  \frac{6\log(T)}{(f^{-1}(\Delta_{\min}))^2} +  2\zeta(3) \Big] k \cdot \Delta_{\max}.$$
\end{restatable}
A detailed proof is provided in
supplementary material.
%Appendix \ref{appndx:missing}. 
Note that the proof technique closely follow Theorem \ref{thm:Regret_CSUCB_lipschitz}.  We remark here that we recover the regret bound of \cite{CHE13} for non-sleeping combinatorial bandits case i.e., when $A_t = [k]$ for all $t$.
Further, observe that when rewards are additive, i.e. $\R_{S_t} = \sum_{i\in S_t} X_{i,t}$ one can achieve $\tilde{O}(\sqrt{T})$ regret bound \cite{kveton2015}; however it is not clear  if the $\tilde{O}(\sqrt{T})$ regret upper bound holds under bounded smoothness assumption. Finally, the instance-independent regret (Theorem \ref{thm:Regret_CSUCB_independent_BoundedMax} and Theorem  \ref{thm:Regret_CSUCB_independent}) guarantee under bounded smoothness condition follows trivially by choosing  $ C = \sup_{x \in[0,1] } f(x)$.

\section{Simulation Results}
\label{sec:simulation}
\begin{figure*}[ht!]
\label{fig: first}
\begin{subfigure}{.47\columnwidth}
\centering
\includegraphics[width=\columnwidth, height=3.7cm]{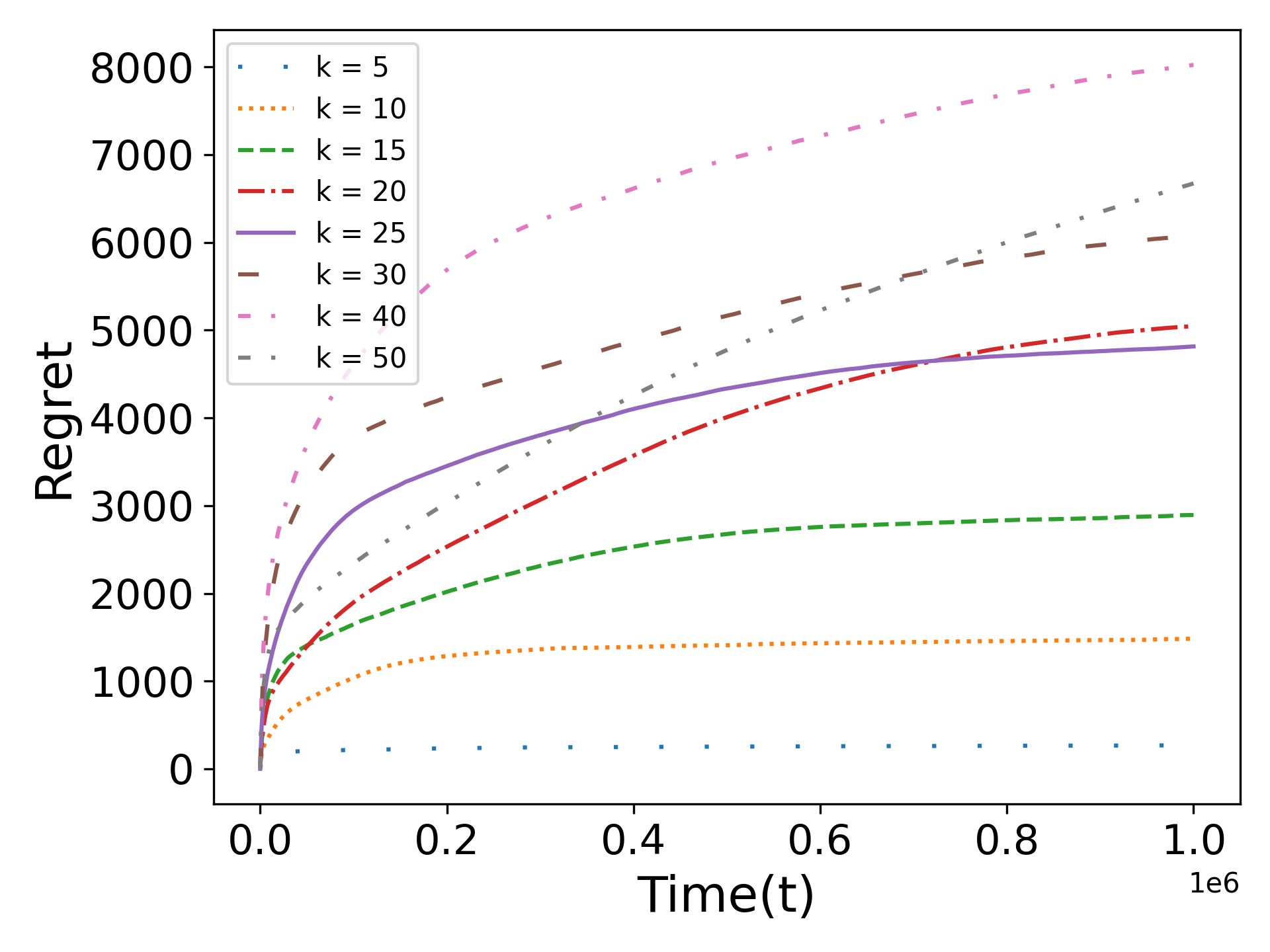}
\end{subfigure}%
      \begin{subfigure}{.5\columnwidth}
      \includegraphics[width=\columnwidth, height=4cm]{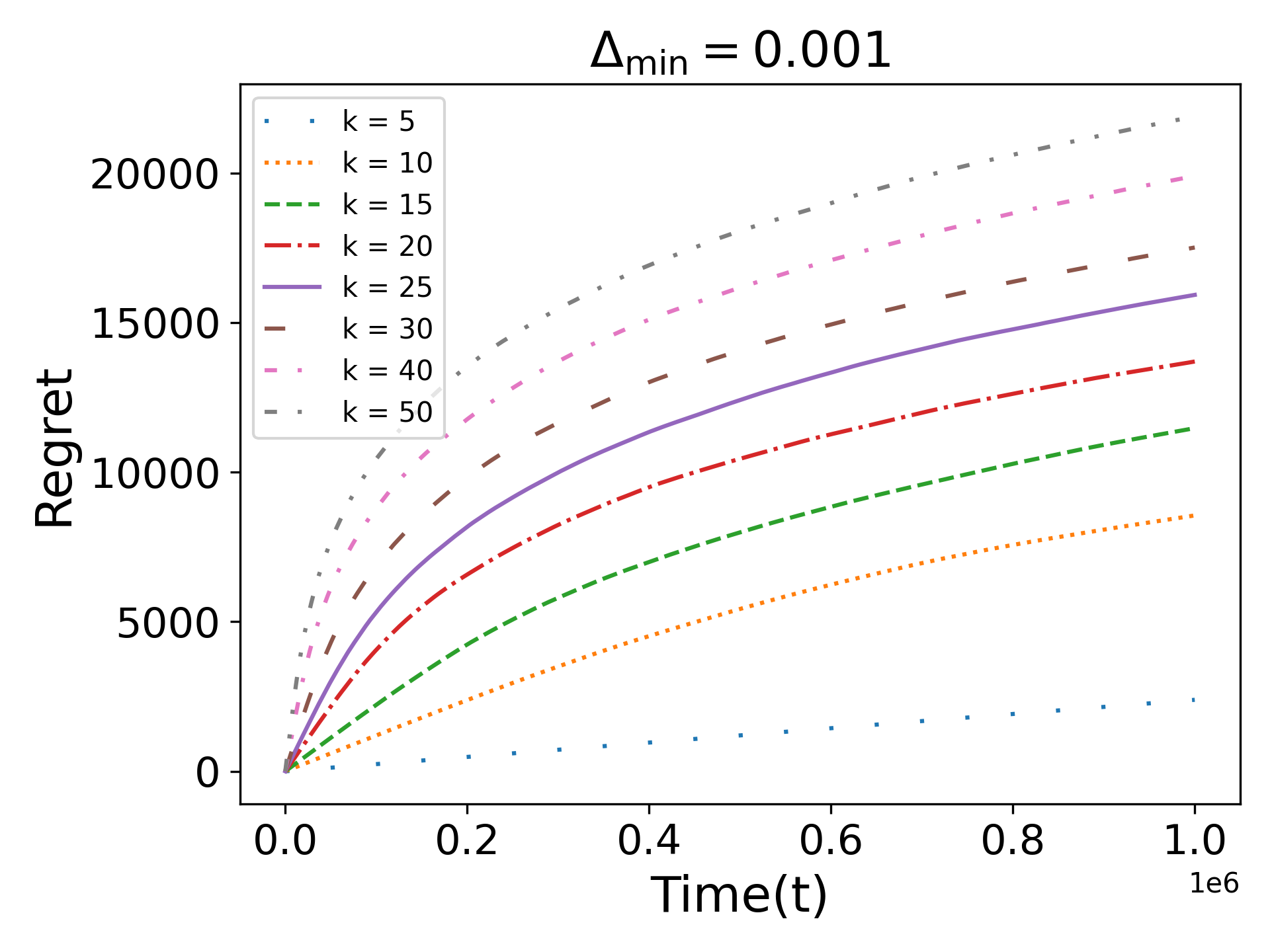}
    \end{subfigure}
    \begin{subfigure}{.5\columnwidth}
    \includegraphics[width=\columnwidth, height=4cm]{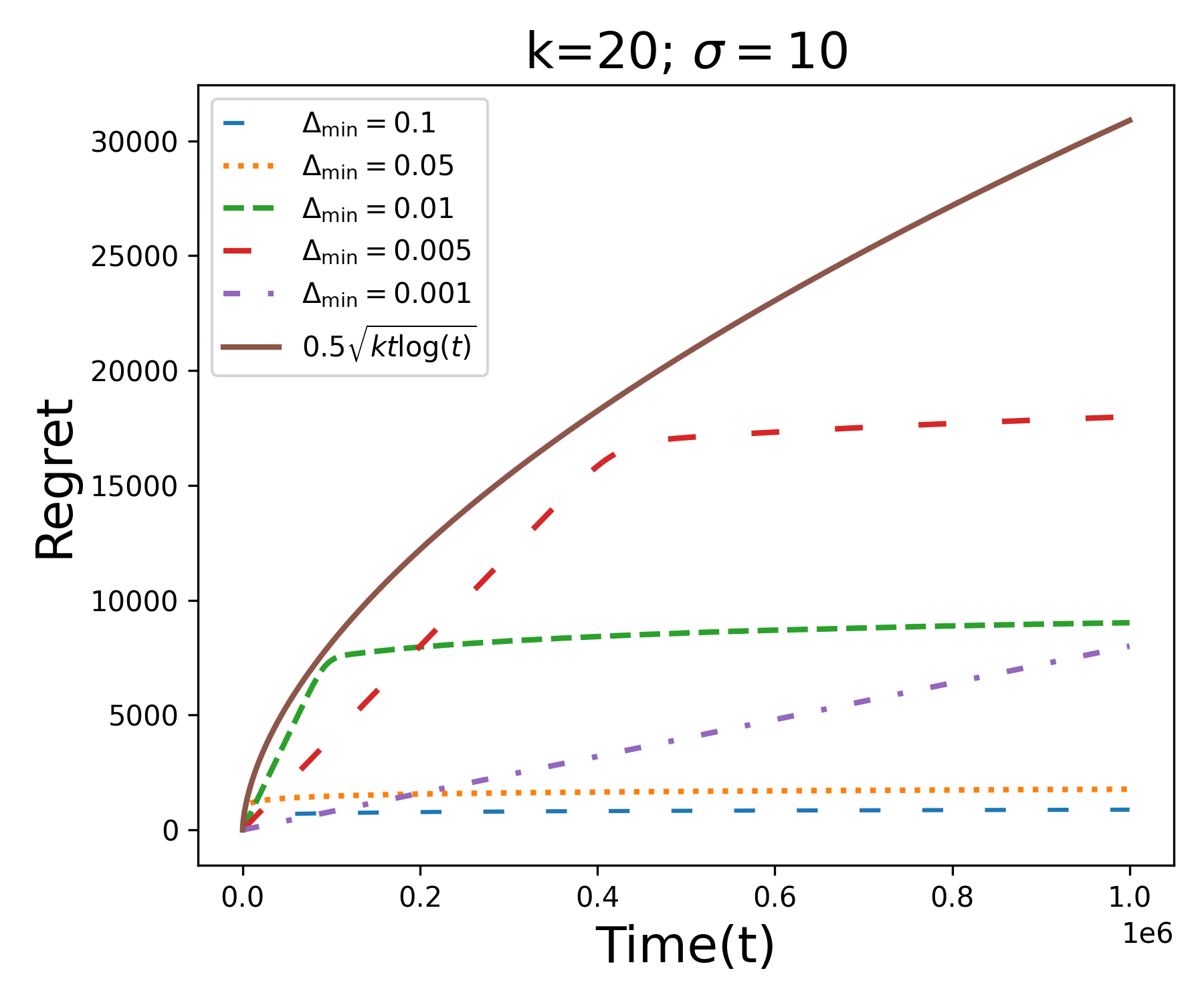}
    \end{subfigure}
     \begin{subfigure}{.5\columnwidth}
        \includegraphics[width=\columnwidth, height=4cm]{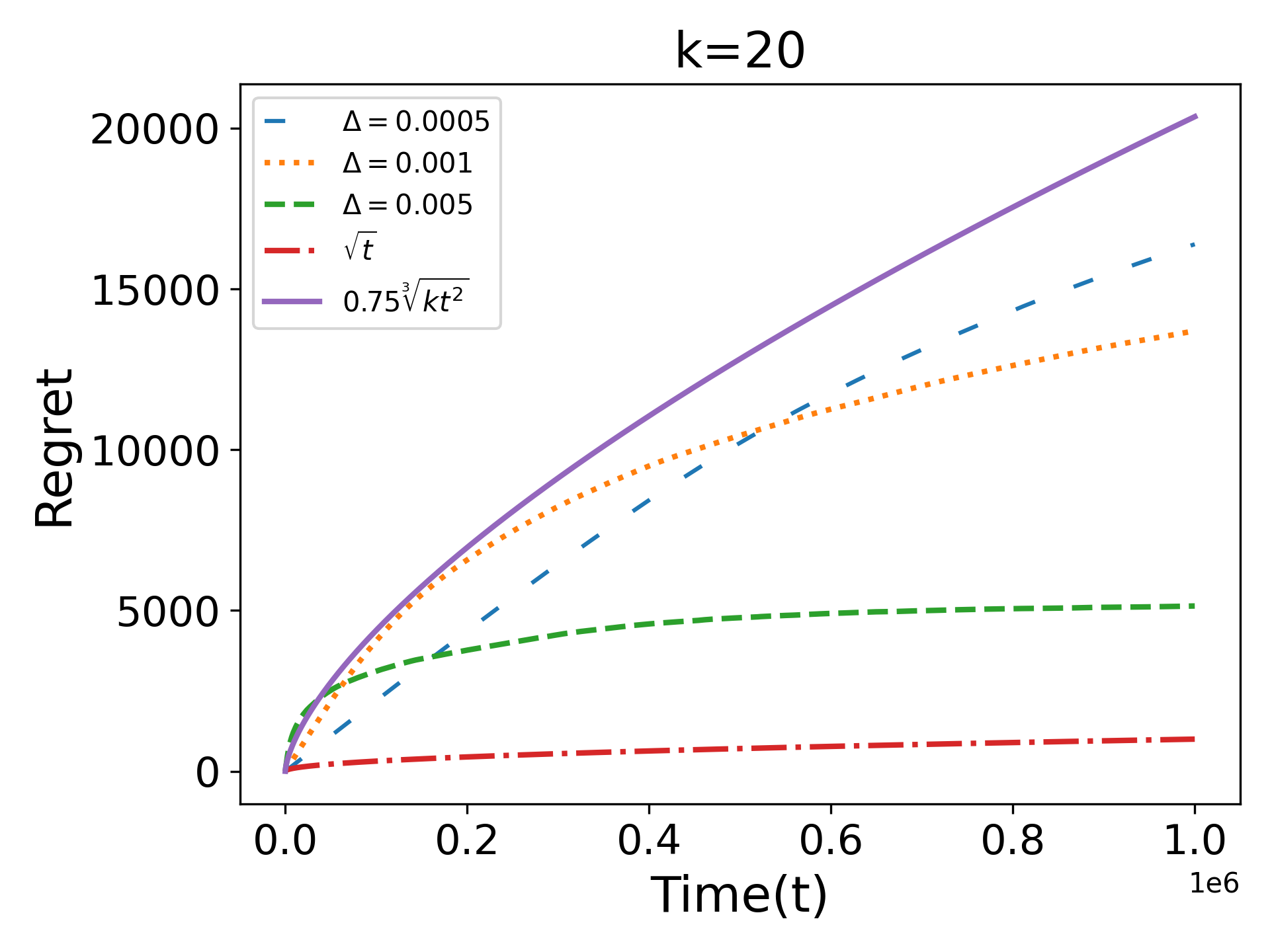}
    \end{subfigure}
    \label{fig:One}
    \caption{Regret Vs Time Plots For $\textsf{UtilReward}$: From L to R, (a) \textsf{ExpOne}: Instance-dependent  regret with randomly generated qualities (Theorem 1) (b) \textsf{ExpOne}: Instance-dependent guarantee for  $\Delta_{\min}$ = 0.001 (c) \textsf{ExpTwo}: Weak instance-dependent guarantee (Theorem 2) (d) \textsf{ExpTwo}: Instance-independent regret guarantee (Theorem 3).}
    \label{fig: first}
\end{figure*}

\begin{figure*}[ht!]
    \label{fig: second}
\centering
\begin{subfigure}{.5\columnwidth}
\includegraphics[width=\columnwidth, height=4cm]{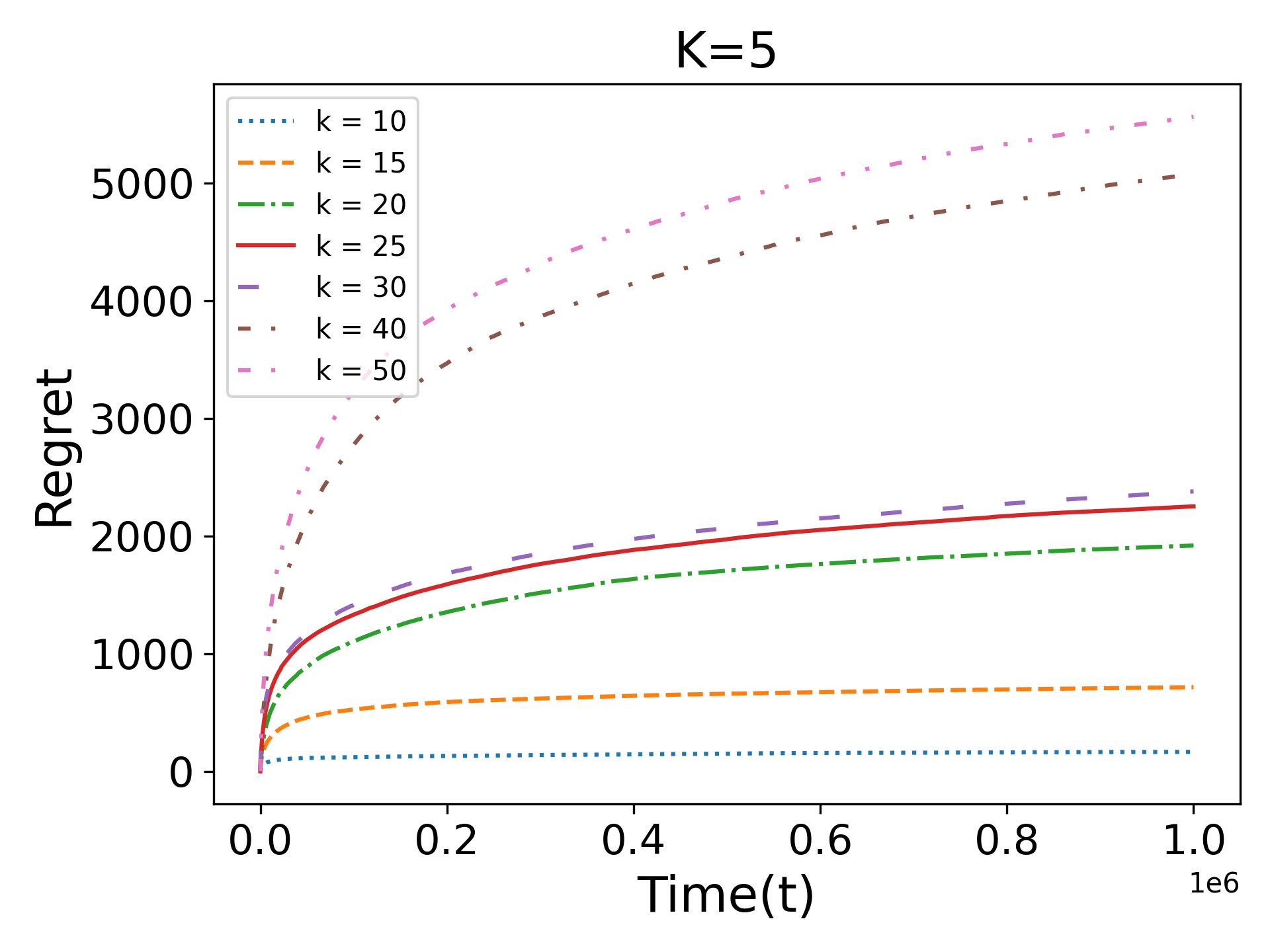}
\end{subfigure}%
        \begin{subfigure}{.5\columnwidth}
        \includegraphics[width=\columnwidth, height=4cm]{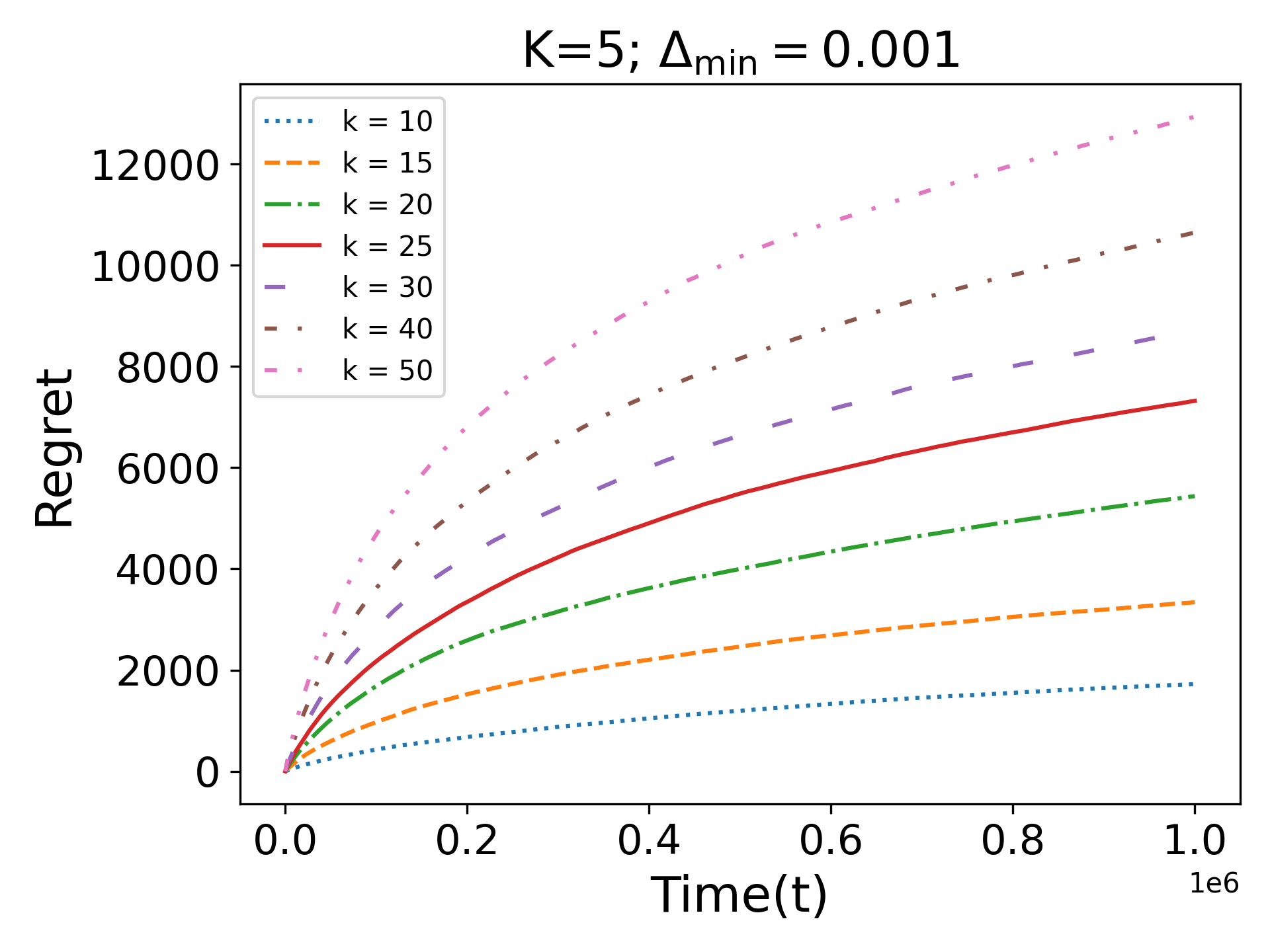}
    \end{subfigure}
    \begin{subfigure}{.5\columnwidth}
    \includegraphics[width=\columnwidth, height=4cm]{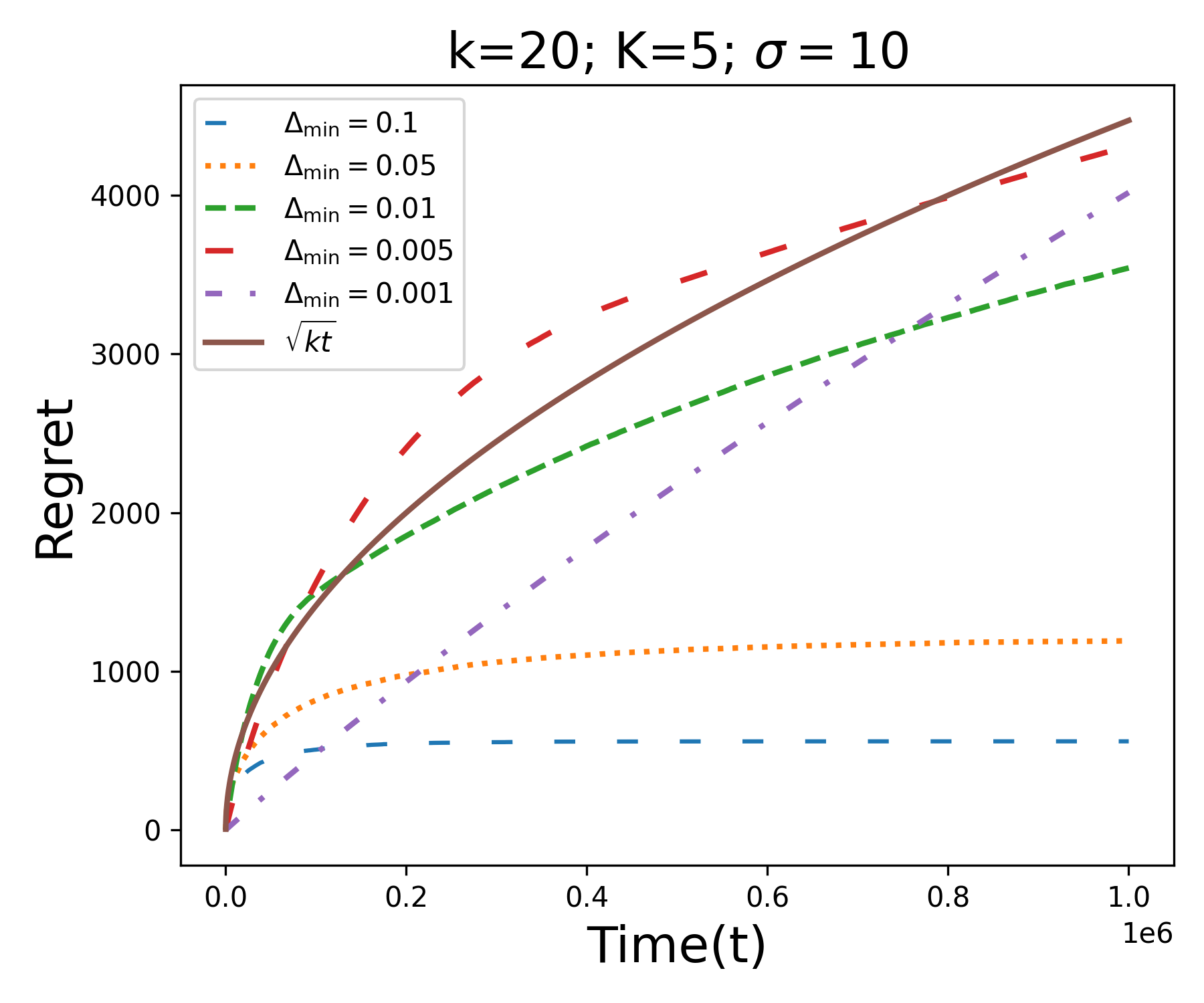}
    \end{subfigure}
        \begin{subfigure}{.5\columnwidth}
        \includegraphics[width=\columnwidth, height=4cm]{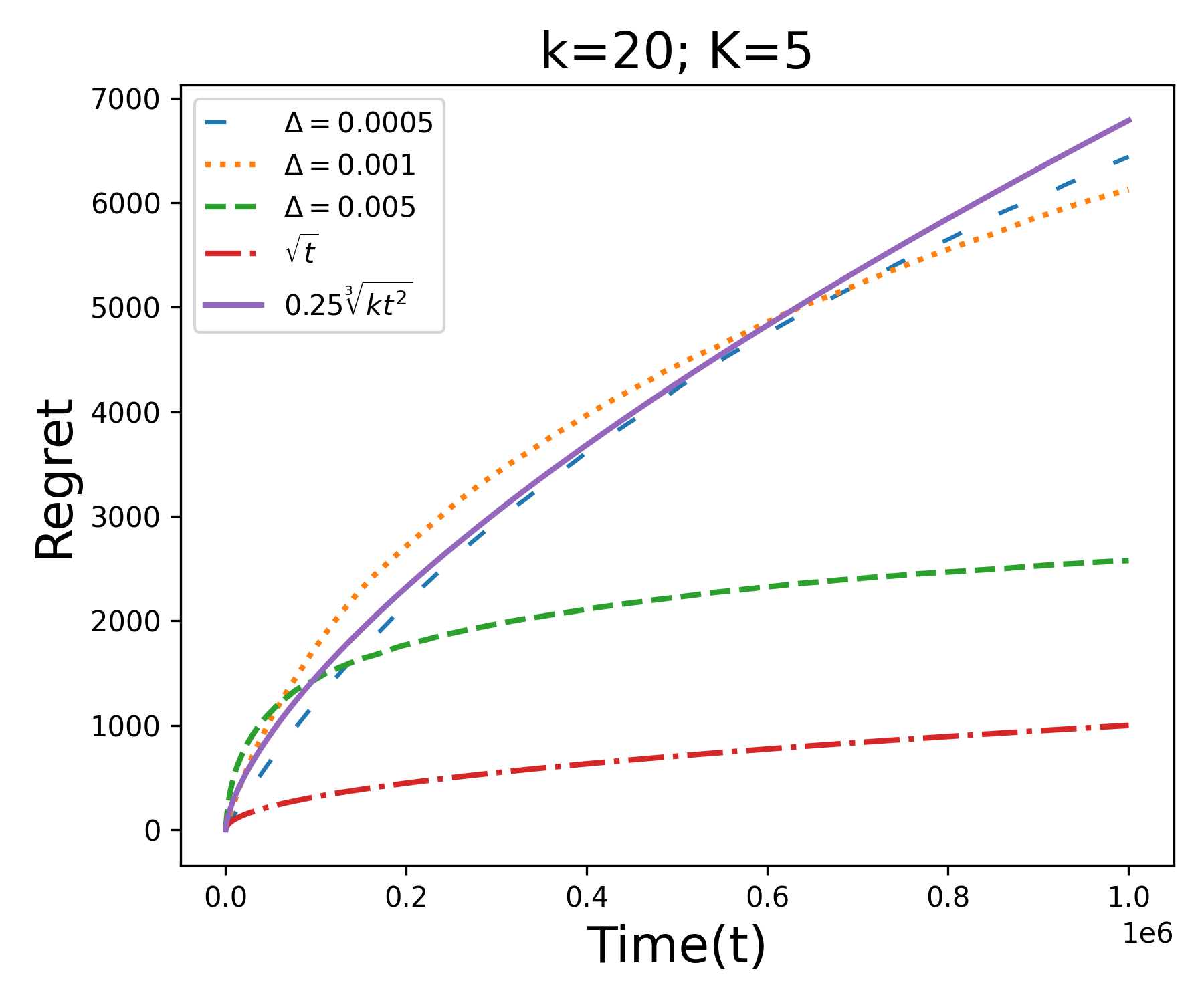}
    \end{subfigure}
    \caption{Regret Vs Time Plots For $\textsf{TopKReward}$: From L to R, (a) \textsf{ExpOne}: Instance-dependent  regret with randomly generated qualities (Theorem 1) (b) \textsf{ExpOne}: Instance-dependent guarantee for  $\Delta_{\min}$ = 0.001 (c) \textsf{ExpTwo}: Weak instance-dependent guarantee (Theorem 2) (d) \textsf{ExpTwo}: Instance-independent regret guarantee (Theorem 3). }
    \label{fig: second}
\end{figure*}
In this section we validate the theoretical results  of the paper using   different reward functions on simulated data.  In particular, we perform experiments on two different  combinatorial bandits settings studied in the literature \cite{jain2018quality,kveton2015combinatorial}. In the first setting, the average quality  of base arm $i$ takes the form $ a_i \cdot \mu_i - b_i$; here $\mu_i$ is a mean of the Bernoulli random variable and $a_i$ and $b_i$ are unknown constants.     In this setting, the quality is also referred to as the utility from arm $i$; where $a_i\cdot X_{i}$ being random reward with mean $\mu_i$ and  $b_i$ being the fixed cost corresponding to arm $i$. We call  this reward setting \footnote{ See \cite{jain2018quality} for detailed motivation and applications of  this  setting.}  as  $\textsf{UtilReward}$.  The goal is to select all the available base  arms with positive quality.  In the second setting (which we call $\textsf{TopKReward}$), we consider the problem of pulling  top $K$ (in terms of quality) available arms and the reward function is additive \footnote{ More details and the regret analysis in \emph{non-sleeping} case is given in \cite{kveton2015combinatorial}.}.  In this setting, note that,  if at some time $t$,  $|A_t| \leq K$,  all the available arms are pulled and the regret at time instant $t$ is zero.  Further observe  that,   both the settings admit polynomial time  exact oracles; i.e. $(1,1)$-\textsc{Oracle}.  
%As there is no previous work which addresses the setting which we have considered, hence we have focused on projecting the performance of our algorithm with respect to our theoretical results. 
%Though we compare the performance of the algorithm when the arms are available at all times, i.e., with CUCB.  
%It is worth noting there may be requirement to design new oracle based on characteristics of the problem being addressed. 
%that the algorithm will work for the problems which falls in our setting, though the design and implementation of the oracle may vary from problem to problem.

\subsection*{Simulation Setup and Observations}
We run two experiments for each of the reward settings mentioned above. In the first experiment which we call $
\textsf{ExpOne}$, the quality parameter $\mu_i$ of each  of the base arms $i$ is chosen independently from uniform distribution over interval $[0.3, 0.8]$. A quality feedback from the base arm $i \in S_t$ is an independent  sample from a Bernoulli distribution with mean $\mu_i$. The availability parameter corresponding to arm $i$ is uniformly sampled from $[0.4, 0.9]$. Similar to the quality  feedback,  availability of $i$ is decided by a random draw from  a Bernoulli distribution with a given availability parameter. 

The second experiment,  $\textsf{ExpTwo}$, is designed to  validate the results of Theorem \ref{thm:Regret_CSUCB_independent_BoundedMax} and \ref{thm:Regret_CSUCB_independent}. The availability of base arms is generated using same approach as in the first experiment. However, the qualities of  base arms $ $ is fixed to be close to each other. We validate the result of Theorem \ref{thm:Regret_CSUCB_independent_BoundedMax},  by fixing the value of $\sigma := \Delta_{\max}/\Delta_{\min}$ and  varying the values of $\Delta_{\min} $, and Theorem \ref{thm:Regret_CSUCB_independent}  by varying the values of $\Delta_{\min}$.  Each of the experiments is executed over time horizon $T=10^6$ and the average rewards from  50 independent runs.
%are plotted in Figures \ref{fig: first} and \ref{fig: second}.

We present the plots associated to $\textsf{UtilReward}$ reward function in Fig. \ref{fig: first}. The first two plots in Fig.  \ref{fig: first} show that as $\Delta_{\min}$ value decreases, the expected regret guarantee of Theorem \ref{thm:Regret_CSUCB_lipschitz} becomes vacuous. The next two plots show that the the regret dependence on time horizon increases from $\sqrt{T}$ to $\sqrt[3]{T^2}$ for similar values of $\Delta_{\min}$ when we fix   $\sigma $ and change $\Delta_{\min}$ to arbitrary values of $\Delta_{\min}$ and $\Delta_{\max}$.  Similar results were observed for different values of $\sigma$, $k$, $\Delta_{min}$ and reward function  $\textsf{TopKReward}$ (Fig. \ref{fig: second}). 
%This behavior is captured graphs in accordance to the expectation from our theoretical results the cumulative regret follows logarithmic trajectory. 
\iffalse
In Fig. \ref{fig:sub3} and \ref{fig:sub4}, we compare the regret when the the true mean of the base arms close is of the order $1/ \sqrt{T}$. This leads  $\Delta_{\min}$ to be very small, which makes the task to distinguish the optimal super-arm from others. In this case, observe that in both reward settings, the regret is sub-linear and roughly is of the order $\sqrt[3]{kT^2}$. In Fig. \ref{fig:sub9} we show the performance of the algorithm with respect to the theoretical result, which validates our theoretical results.

In Fig. \ref{fig:sub5} and \ref{fig:sub6}, we compare the algorithm with CUCB (i.e., when all arms are available), when the reward function is to select arms with positive rewards. Similarly, in Fig. \ref{fig:sub7} and \ref{fig:sub8}, we compare the algorithm with CUCB, when the reward function is to select top K arms. From our simulations we show that our algorithm is effective and it performs in agreement with our theoretical results.
\fi
\section{Related Work}
\label{sec:reltd_wrk}
The stochastic bandits problem has been extensively studied in the literature  \cite{lai1985,auer2002,thompson1933likelihood,agrawal12}.  We refer the reader to \cite{lattimore2018bandit,slivkins2019} for a book exposition on multi-armed bandits and their applications. Most previous work in literature--- with few exceptions such as \cite{CHA17,KLE10,li2019combinatorial} --- assume that all the arms are available at all time instants. It is shown that the classical algorithms,  adapted appropriately, are also optimal in a sleeping bandits setting \cite{CHA17,KLE10}.    %\cite{KLE10} study the sleeping MAB setup where an arbitrary set of arms is available at any time and define sleeping regret as a performance metric to evaluate algorithms. Further, they propose AUER, an  UCB-style algorithm that attains a logarithmic regret guarantee. In \cite{CHA17} the authors show that the Thompson sampling-based Bayesian counterpart of AUER also achieves logarithmic sleeping regret. 

Combinatorial multi-armed bandits (CMAB) is another well studied  variant of stochastic MAB problem which considers multi-pull setup \cite{bianchi12cmab,shou_NIPS2014,CHE13,combes_NIPS2015,gai10,gai12,kveton2015,li2019combinatorial,ontanon2013combinatorial,WANG18,wen2015}.
%Many papers have contributed to this literature by analyzing the regret with different reward functions such as additive, stochastic, adversarial, etc. under different feedback structures such as bandit and semi-bandit feedback. \cite{kveton2015,gai10,gai12} consider additive stochastic CMAB with semi-bandit feedback, whereas \cite{bianchi12cmab} considers additive adversarial CMAB with semi-bandit feedback . 
%\cite{shou_NIPS2014} considers CMAB setting where dedicated exploration is needed, and the objective is different from minimizing the expected regret. \cite{combes_NIPS2015} considers additive CMAB with both stochastic and adversarial rewards as well as semi-bandit and bandit feedback, and provide problem-specific regret bounds.
%\cite{gopalan2014thompson}: bandit feedback, inter-dependent super-arms; 
%Except for a few works such as \cite{CHE13,gopalan2014thompson}, most of the papers address additive rewards settings. 
%\cite{gopalan2014thompson} considers general CMAB with bandit feedback, with inter-dependent rewards.
% \cite{kveton2015} considers the combinatorial bandits with additive rewards; i.e., the reward of any set of arms is a sum of the rewards of its constituent arms. They show that their proposed algorithm, \textsc{CombUCB1},  achieves logarithmic instance-dependent regret and $O(\sqrt{T \log(T)})$ instance-independent regret upper bound. 
\cite{CHE13} consider a  general reward function with some smoothness condition and proposed CUCB, a UCB-style algorithm.   In contrast, we consider arbitrary arm availability and general rewards and show that CUCB when extended to sleeping bandits setting achieves optimal regret guarantee.  
 We also remark here that their analysis does not generalize to combinatorial sleeping bandits, and hence we need novel proof techniques to bound the sleeping regret in a CMAB setting. To the best of our knowledge, we are the first to address combinatorial sleeping MAB with a general reward structure and provide instance-dependent as well as instance-independent regret upper bound.

The closest work to this work is \cite{chen2018contextual}.
Similar to their work we consider semi-bandit feedback and combinatorial sleeping bandits framework. However, \cite{chen2018contextual} considers contextual bandits setting, whereas we study a  sleeping combinatorial  MAB setting. The proposed algorithms (CC-MAB and CS-MAB, respectively) differ crucially in how they carry out exploration. CC-MAB explores the subset of available arms if it contains at least a single unsaturated base arm (\cite{chen2018contextual}, Algorithm 2, Line 7). Hence, the exploitation is carried only if all the available base arms are saturated. In contrast, CS-UCB  does not demarcate the exploration and exploitation in  this manner. So, even if some ``obviously" bad super-arms are not explored, CS-UCB does not pull them. Also, there are following two important differences in the setting considered. Firstly, \cite{chen2018contextual} consider that the reward function is submodular, whereas we consider general reward functions. Indeed, if the reward function satisfies submodularity, our results can be extended easily by considering $(1-\frac{1}{e})$- approximation oracle. Secondly, they consider that the time horizon is a-priori known to the algorithm, which may be an unrealistic assumption in many practical cases. Note that we provide an any-time regret guarantee, i.e., $T$ is not given as an input to the algorithm. Also, they proved that CC-MAB achieves the regret of $O(\sqrt[3]{T^2 \log(T)})$ for the specific case of the submodular reward function, whereas we provide similar  regret bound with more general reward functions. The recent work of \cite{NIKA20} also studies contextual combinatorial bandits set up with sleeping arms and semi-bandit feedback. The authors consider a setting where the arms are differentiated based on the context.

\section{Conclusion and Future Work}
\label{sec:conclusion}
In this paper, we considered combinatorial sleeping multi-armed bandits setting where a subset of arms is available at a given time instant. We  analyzed the \CSUCB\ algorithm and analyzed its regret guarantee under two setups; \LS\ and \BS. We showed that under \LS\ setting, \CSUCB\ achieves $O(\log (T)/\Delta_{\min})$ instance-dependent sleeping regret guarantee. Additionally, we prove that \CSUCB\ achieves $O(\sqrt{T \log(T)})$ weak instance-dependent regret under the assumption that the ratio of maximum and minimum achievable rewards is bounded. Also,  we provide $O(\sqrt[3]{T^2 \log(T)})$ instance-independent regret in the most general case. We also show that \CSUCB\ in \BS\ setting matches the conventional regret guarantee for the combinatorial MAB setting under the same set of assumptions (i.e., $O(\log(T))$). Finally, we validate the proven theoretical guarantees through experiments.

The instance-independent regret guarantee under \BS\ setting remains an interesting open problem.  Also, a finely tuned analysis with availability specific regret guarantees is an interesting future direction. This setup could be used together with other MAB settings, for instance, rotting bandits \cite{levine2017rotting}, where the arm pulling strategy may lead to the dropping of the arms.

%\newpage 

\bibliographystyle{alpha}
\bibliography{references}
\onecolumn
\appendix
\section{ Preliminaries}
\begin{definition}
Let $X_1, X_2,.....,  X_n$ be $n$ independent random variables, and $S_n = X_1 + X_2 +....+ X_n$, where $\forall i, X_i \in [a_i ,b_i]$, then according to Hoeffding's inequality,
 \begin{center}
 $\mathbbm{P} \{ S_n - \mathbbm{E}[S_n] \geq t\} \leq e^{\frac{-2t^2}{\sum_i (b_i - a_i)^2}}$
 \end{center}
 \end{definition}

\subsection{Notation}
\label{appndx:notation}
\begin{table}[H]
  \centering
    \label{table:notation}
  \begin{tabular}{|l|l|}
      \hline
      $[k]$ & Set of arms. \\
      
      $T$ & Time horizon (or the number time steps). \\
      
      $A_t$ & Set of arms available at time $t$. \\ 
      
      $\mu_i$ & Bernoulli parameter (or mean) for arm $i$.
      \\
      
      $\muvector_t$ & The vector of mean rewards of each arm in set $A_t \subset [k]$ at time $t$. 
      \\ 
      
      $\overline{\muvector}_t$ & The vector of UCB estimates of means (unknown) for each arm in set $A_t \subset [k]$ at time $t$
      \\ 
      
      $S_t $ & The subset of arms (super-arm) pulled at time $t$.
         \\
    
    $\R_{S_t}$ & Reward obtained when super-arm $S_t$ is pulled at time $t$. 
      \\ 
      $S^{\star}_t $  & $ \arg \max_{S \subseteq A_t} R_S$ 
      \\ 
      
            $S_{B}(A)$ & The set of all bad super-arms  i.e. $S_{B}(A) = \{S \subseteq A | \Delta_{S} > 0   \}$. 
    \\
    
      $X_{i,t}$ & A random reward obtained at time $t$ from arm $i$.
      \\ 
          
          $N_{i,t}$ & Number of times arm $i$ is pulled till $t$ time steps.
      \\
      
      $\hat{\mu}_{i,t}$ &  $X_{i, 1:t}/N_{i,t}$; Empirical estimate of arm $i$ till time step $t$.
      \\
      
      $\varepsilon_{i,t}$ &  $\sqrt{3\log (t)/2 N_{i, t}}$; UCB confidence interval of arm $i$ till step $t$.
      \\
      
      $\varepsilon_{t}$ &  $\sqrt{3\log (t)/2 \ell_t}$.
      \\
      
%      $\Delta_{i,j}$ & $| \mu_i - \mu_j |$.
%      \\
%      $\ell_ t$ & The minimum number of time steps an available arm is pulled till round $t$. \\ 
      %$\ell_t$ &  $\frac{6 \log(t)}{(f^{-1}(\Delta_{\min}))^2}$ \\ 
      %\hline
      %$\ell$ &  $\frac{6 \log(T)}{(f^{-1}(\Delta_{\min}))^2}$\\
      %\hline
      $T_e$ & The set of time instances $t$ such that $A_t$ is explored, i.e., all arms in $A_t$ are saturated.\\
      
      $T_u$ &  $[T] \setminus T_e$.\\
       
      $A_{e,t}$ & The set of saturated arms that are available at time $t$.\\
      
      $A_{u,t}$ & $ A_{t} \setminus A_{e,t}$ \\
      
          $\Delta_{S_t}$ & $ \gamma \cdot \R_{S_t^{\star}}(\boldsymbol{\mu}) - \R_{S_t}(\muvector) $; Quantitative measure for sub-optimality of super-arm $S_t$.\\ 
    \hline 
%      $w_{i,t}$ & Number of times arm $i$ is optimal until time $t$.
%      \\
%      $e_{i,t}$ & $\sqrt{32 \ln(w_{i,t})/n_{i,t}}$.
%      \\
%      $Q_{i,j}$ & $32 \ln(w_{i,t}) / \delta_{i,j}^2$.
%      \\
    \end{tabular}
        \caption{Notation Table}
  \end{table}

\subsection{Algorithm: CS-UCB}
\begin{algorithm}[ht] 
\caption{\CSUCB}
\label{algo:cs-ucb}
\begin{algorithmic}[1]
\STATE Initialization:
\FOR{$i \in [k]$}
\STATE $N_{i,0} = 0$, $\overline{\muvector}_{i,0} = 1$, $X_{i,0}=0$ \;
\ENDFOR
\FOR{$t = 1,2,3,\ldots $}
 \STATE Observe set of available arms as $A_t$ \;
\IF{$\exists j \in A_t$, such that, $N_{j,t} = 0$}
\STATE Select $S_t = A_t$ \;

\ELSE
\STATE $S_t = \textsc{Oracle}(A_t,  
\overline{\muvector}_t)$ \;
\ENDIF
\STATE \textbf{Observe:} Semi-bandit feedback as $X_{j,t} \in \{0,1\}, \forall j \in S_t$ \ and $\R_{S_t}(\muvector)$;
\STATE \textbf{Update:} 
\begin{multicols}{2}
\begin{itemize}
    \item  $N_{i,t} = \begin{cases}   N_{i,t-1}  &  \text { if } \forall i \notin S_t  \\   N_{i,t-1} +1  &  \text { if } \forall i \in S_t  \end{cases}$ 
    \item $X_{i, 1:t} = \left\{ \begin{array} { l l } { X_{i,1:t-1} } & { \text { if } \forall i \notin S_t } \\ {X_{i, 1:t-1} +X_{i,t} } & { \text { if } \forall i \in S_t} \end{array} \right.$ 
\item $\overline{\mu}_{i, t} =  \frac{X_{i, 1:t}}{N_{i,t}} + \sqrt{\frac{3\log (t)}{2 N_{i, t}}} $ 
\item [ ]
\end{itemize}
\end{multicols}
\ENDFOR
\end{algorithmic}
\end{algorithm}
%\ganesh{We need to explain the algorithm in detail here.}

%\sg{this para of talkig about esults is repetiative as in section 5 we say the same with more explaination.}In our first result (Theorem \ref{thm:Regret_CSUCB_lipschitz}), we show that --- under \LS\ setting --- \CSUCB\  attains asymptotically optimal (up-to  constants) instance-dependent sleeping regret upper bound.  We also show similar upper bound under \BS\ setting (Theorem \ref{thm:Regret_CSUCB1_bounded}). Finally, in Theorem \ref{thm:Regret_CSUCB_independent_BoundedMax} we prove instance-independent sleeping regret bound of ($O(\sqrt{T\log T})$) under a mild assumption and in Theorem \ref{thm:Regret_CSUCB_independent} ($O(\sqrt[3]{T^{2}\log T})$) in general under \LS\ setting. 
\newpage
\section{Omitted Proofs}
We begin with introducing additional notation used in rest of the paper. Let $R_{S}(\overline{\muvector} ; \muvector ) $ denote the reward obtained  according to the quality vector $\muvector$ from the set $S$ which also satisfies the condition that   $\R_{S}(\overline{\muvector}) \geq \gamma \cdot \R_{S^{\star}}(\overline{\muvector}) $. Here, $\R_{S^{\star}} =  \max_{S \in A} \R_{S}(\overline{\muvector}).$  
\label{appndx:missing}
\ObservationOne*
\begin{proof}
Using Hoeffding's lemma \cite{hoeffding1963} we have,  
\begin{equation}
\label{eqn: hoffding type bound}
    \mathbbm{P} \{|\hat{\mu}_{i,t} - \mu_i| \geq  \varepsilon_{t}\}  \leq 2 e^{-2 N_{i,t} \varepsilon_{t}^2}  = 2 e^{-3 N_{i,t} \log(t)/ \ell_t} \leq 2/t^3.
\end{equation}
Here, $\hat{\mu}_{i,t}$ is the empirical mean  reward of arm $i$ till time $t$. 
The last inequality follows from the fact that $N_{i,t} \geq \ell_t$.  Further, 
 from the definition of $\overline{\muvector}$,  with probability at least $1 -\frac{2}{t^3}$, 
\begin{equation}
\label{eq:8}
\varepsilon_{t} \underset{(i)}{ >}    | \hat{\mu}_{i,t} - \mu_i |  \underset{(ii)}{ =}  | \overline{\mu}_{i,t} - \mu_i  - \varepsilon_{t}| \underset{(iii)}{ \geq } | \overline{\mu}_{t} - \mu_i | - \varepsilon_{t}. 
\end{equation}

Here, $(i)$ follows  from Equation \ref{eqn: hoffding type bound}, $(ii)$ is immediate from the definition of $\muvector$ and finally $(iii)$ follows from the  triangle inequality.  Thus, we have $| \overline{\mu}_{i,t} - \mu_i | < 2 \varepsilon_{t} $ with probability atleast $1 - 2|S|/t^3$.
%Hence, we have that  $ \mathbbm{P} \{ \max_{i \in S} |\hat{\mu}_{i,t} - \mu_i| < 2 \varepsilon_{t} \} \geq 1 - \frac{2|S|}{t^3} $.
\end{proof}

\ObservationTwo*
\begin{proof}
The monotonicity property  and Lipschitz smoothness implies that, 
\begin{align}
    | \R_{S_t}(\overline{\muvector}_t) - \R_{S_t}(\muvector) | & =  \R_{S_t}(\overline{\muvector}) - \R_{S_t}(\muvector) \leq C \max_{i \in S_t} |\overline{\mu}_{i,t} - \mu_i| \tag{Monotonicity property and Lipschitz property}   \\
    \intertext{However, }
    \R_{S_t}(\overline{\muvector}_t) - \R_{S_t}(\muvector) & \geq \gamma \cdot  \R_{S_{t}^{\star}}(\overline{\muvector}_t) - \R_{S_t}(\muvector_t)  \geq \gamma \cdot \R_{S_t^{\star}}(\muvector_t) - \R_{S_t}(\muvector) = \Delta_{S_t}.
\end{align}
Further, from the definition of $\overline{\mu}_{i,t} = \hat{\mu}_{i,t} + \varepsilon_{i,t}$, where $\varepsilon_{i,t} = \sqrt{3\log(T)/2 N_{i,t}}$ for arm $i$ till time $t$, we have \begin{equation} |\overline{\mu}_{i,t} - \mu_i| \leq |\hat{\mu}_{i,t} - \mu_i | + \varepsilon_{i,t}  \leq 1 + \sqrt{3\log(T)/2}.\end{equation} 
From Eq. 3 and Eq. 4, observe that $\Delta_{S_t} \leq C \max_{i \in S_t} |\overline{\mu}_{i,t} - \mu_i| \leq C (1 + \sqrt{3\log(T)/2}) $.
\end{proof}
%%%%%%%%%%%%%%%%%%%%%%%%%%%%%%%%%%%%%%%%%%%%%%%%%%%%%%%%%%%%%%%%%%%%%%%%%%%%%%%%%%%%%%%%%%%
\iffalse
\RegretLipschitzDependent*
We choose $\ell_t = \frac{6C^2\log(t)}{\Delta_{\min}^2}$ and $\varepsilon_t = \sqrt{\frac{3 \log(t)}{2 \ell_t}} $ and divide the time instants into sets  $T_e$ and $T_u$ as described in Section \ref{sec:our_algo}. 
We organize the proof of the theorem in the following steps. 

\begin{itemize}[noitemsep, nolistsep]

    \item[1] We first show that the probability that a sub-optimal super-arm is pulled at any time $t \in T_e$ is very small and hence the expected sleeping regret incurred in these time steps is small.
    \item[2] Next, we show that the number of time steps $t$ such that  $t \in \{ t^{'}| t^{'} \in T_u \text{ and } \exists i \in S_t  \text{ such that } N_{i,t} < \ell_t \} $ is upper bounded by $k\ell_T$.
    \item[3] In the last step, we consider the time instances $t \in T_u$ such that all the pulled arms $i \in S_t$ are saturated. In this case, we show that with high probability, we have $S_t \notin S_{B}(A_t)$.  
\end{itemize} 
\textcolor{red}{Elaborate these steps more and add the proof there itself. Looks like one has to go through the proof twice... }

\noindent \underline{\textbf{Step 1}}: In this step, we consider time steps $t \in T_e$. We begin with the following supporting lemma. 
Recall that $B_t$ is an event that the oracle returns $\gamma$-approximate solution i.e. $r_{\overline{\muvector}_t}(S_t) \geq \gamma \cdot r_{\overline{\muvector}_t}(S^{\star}(A_t))$.
\fi

\ExploredUpperBoundLS*
\begin{proof}
Let $S_{t}^{\star} = \arg\max_{S \subseteq A_t} \R_S$ and the event $B_t$ has occurred. We prove the lemma using the following supporting claim. 
\begin{claim} 
\label{clm:supportclm}
Let $t \in T_e$ and  $S_t = \textsc{Oracle}(A_t,  
\overline{\muvector})$ and $S_t^{'} = \textsc{Oracle}(A_t, \muvector)$. Then
    $\mathbbm{P}\{ \R_{S_t}(\muvector) = \R_{S_t^{'}} (\muvector) \}  \geq 1 - 2|A_t|/t^3$.
\end{claim}
To see the proof of the lemma  observe that 
%with probability atleast $1 - \frac{2|A_t|}{t^3}$ we have,  
\begin{align*}
    \R_{S_t}(\muvector) &=\R_{S_t^{'}}(\muvector)  \geq \gamma \cdot  \R_{S_t^{\star}}(\muvector)  
    = \gamma \cdot \textsc{Opt}_{\muvector}(A_t).
\end{align*}
The first equality in the above equation is true with probability atleast $1 - 2|A_t|/t^3$ from Claim \ref{clm:supportclm}. The first inequality holds from the fact that the event $B_t$ has occurred.  Hence, we have    $ \mathbbm{P}(S_t \notin S_{B}(A_t)| B_t) \geq 1 - 2|A_t|/t^3.$ This completes the proof of the lemma.
\end{proof}
%\textcolor{red}{Many reviewers had problem with this claim... We need to rewrite the proof of entire lemma in fact. Also the statement of lemma can be modified slightly to  $\mathbbm{P}(S_t \in OPT_{\mu_t}(A_t))$  }

\begin{proof}[Proof of Claim \ref{clm:supportclm}]

First note that, it is enough to  show that $S_t^{'} = S_t$. However, these sets might not be unique and hence we assume $S_t^{'} \neq S_t$.  
Let  $Q_{t}^{\star} \in \arg \max_{S \in A_t} R_{S}(\muvector) $ and   $S_{t}^{\star} \in \argmax_{S \in A_t} R_{S}(\overline{\muvector}) $. 
From the monotonicity property of $\R$ and the definition of  $\overline{\muvector}$ it holds that \begin{equation}
    \R_{S_t}( \overline{\muvector}_t) \geq \R_{S_t^{'}}( \overline{\muvector}_t) \geq  \R_{S_t^{'}}( \muvector) 
    \label{eqn:claim one first inequality}
\end{equation} 
Here, the first inequality follows from the optimality of $S_t$ with respect to $\overline{\muvector}_{t}$ and the  second inequality follows from the monotonicity property. 
For contradiction, let us assume that $S_t \neq S_t^{'}$  and $ \R_{S_t^{'}}(\muvector) >  \R_{S_t}( \muvector) $. Using this   inequality with Equation \ref{eqn:claim one first inequality}  we get $ \R_{S_t}(\overline{\muvector}_{t}) > \R_{S_t}(\muvector) $. From  Lipschitz property we have, 
\begin{equation}
    \R_{S_t }( \overline{\muvector}_t) - \R_{ S_t}(\muvector) = |\R_{S_t }( \overline{\muvector}_t) - \R_{ S_t}(\muvector) | \leq C \max_{i \in S_t } |\overline{\mu}_{i,t} - \mu_i |.
\end{equation}
Let $\ell_t = 6C^2 \log(t)/\Delta_{\min}^2$. As $t \in T_e$, we have $N_{i,t} \geq \ell_t$ for all $i \in A_t$. Hence, from Observation \ref{obs:One},  with probability atleast $1 - \frac{2|A_t|}{t^3}$, we have,  
$\max_{i \in S_t } |\overline{\mu}_{i,t} - \mu_i | \leq \max_{i \in A_t} |\overline{\mu}_{i,t} - \mu_i | < 2 \varepsilon_t $.  This gives, $\R_{S_t }(\overline{\muvector}_{t}) - \R_{S_t }(\muvector)   < 2 C \cdot  \varepsilon_t = \Delta_{\min} $. To see the last inequality recall from Observation \ref{obs:One} that $\varepsilon_t = \sqrt{ \frac{3 \log(t)}{2 \ell_t}}$. 
\iffalse
Consequently,
\begin{align*}
\R_{S_t }(\muvector) + \Delta_{\min} & > \R_{S_t }( \overline{\muvector}_{t}) 
\end{align*}
Here, the last inequality follows from the monotonicity property. 
\fi 
Hence, we have $
 \Delta_{\min} > \gamma \cdot \R_{S_t^{\star}}(\overline{\muvector}) - \R_{S_t}(\muvector) \geq \gamma \cdot \R_{S_t^{\star}}(\muvector) - \R_{S_t}(\muvector) $.
This contradicts the definition of $\Delta_{\min}$. Thus, with probability atleast $1 - \frac{2|A_t|}{t^3}$ we have that $\R_{S_t}(\overline{\muvector}_t) = \R_{S_t}(\muvector_t) $. This completes the proof of the claim.
\iffalse
Note that, $\mathbbm{P}( S_t \in S_{B}(A_t) | B_t) = \mathbbm{P}(  r_{\muvector}(S_t) \leq \gamma \cdot \textsc{Opt}_{\muvector}(A_t)  ) $. First, observe that $r_{\overline{\muvector} }(S_t) >  r_{\muvector} (S^{\star}(A_t))  $
Note that $t \in T_e$ we have that 
\begin{align}
\label{eq:bound_lipschitz}
     |r_{\overline{\muvector}(A_t)}(S_t) - r_{\muvector(A_t)}(S_t)| & \leq   C\cdot \max_{i\in A_t}|\overline{\mu}_{i,t} - \mu_i| \tag{Lipschitz property} \nonumber \\
     \intertext{ Using  $\ell_t = \frac{6C^2\log(T)}{\Delta_{\min}^2 }$ in Observation \ref{obs:One}, we have }
      \mathbbm{P}(|r_{\overline{\muvector}(A_t)}(S_t) - & r_{\muvector(A_t)}(S_t)| <  2C \varepsilon_t ) \geq 1 - \frac{2|A_t|}{t^3}.   \nonumber \\
      \intertext{
      Hence, with probability atleast $1 - \frac{2|A_t|}{t^3}$ we have }
    % & = 2C \sqrt{\frac{3\ln (t+1)}{2 N_{i, t+1}}} \nonumber \\
    |r_{\overline{\muvector}(A_t)}(S_t) -  r_{\muvector(A_t)}(S_t)| & < 2C \sqrt{\frac{3\log (t)}{2 \ell_t}} %\nonumber \tag{$N_{i,t}\geq \ell_t$}  
     = \Delta_{\min}  \nonumber\\ 
     \implies  r_{\muvector(A_t)}(S_t)  +   \Delta_{\min} 
    & > r_{\overline{\muvector}(A_t)}(S_t) \nonumber\\
   &\geq \gamma \cdot r_{\overline{\muvector}(A_t)}(S^{\star}(A_t)) \tag{As $S_t$ is optimal super-arm for $\overline{\muvector}$ and $\gamma \in (0,1]$ } \nonumber \\ 
   & \geq \gamma \cdot r_{\muvector(A_t)}(S^{\star}(A_t))  \tag{from the monotonicity property} \\
   & =  \gamma \cdot \textsc{Opt}_{\muvector}(A_t).  \nonumber  \label{eq:lipstz_contradict} 
\end{align}
\fi
\end{proof}

\LemmaCardinalityOfD*
\begin{proof}
Recall that by definition, we have $|D| := |\{t: \exists j \text{ such that } j \in A_{u,t}, j\in S_t  \} |. $ Hence we have, 
 \begin{align*}
      |D| &=  |\{t: \exists j \text{ such that } N_{j,T} < \ell_T   \} | \leq \sum_{j =1}^{k} |\{t:  N_{j,T} \leq \ell_T  \} | 
      \leq k\ell_T  . \label{eq:step2}
 \end{align*}
 \end{proof}

\TheoremOneLemmaThree*

\begin{proof}
Consider $t \in E$ and recall from Lemma \ref{lem:lemmaOne} that $Q_{t}^{\star}  = \argmax_{S \in A_t} \R_{S}(\muvector)$. 
We have,
\begin{equation*}
    \centering
    \R_{S_t}(\overline{\muvector}_t) \geq \gamma \cdot  \R_{S}(\overline{\muvector}_t^{\star}), \hspace{10pt} \forall S \subseteq A_t .
\end{equation*}
 For all $j \in S_t$  we have $N_{j,t} \geq  \ell_t$. Hence from Observation \ref{obs:One}, with probability atleast $ 1 - \frac{2|S_t|}{t^3}$  
 we get, 
\begin{equation}
    \max_{j \in S_t}|\overline{\mu}_{j,t} - \mu_j| < 2\varepsilon_{t}.
\end{equation}
This implies, 
\begin{align*}
    |\R_{S_t}(\overline{\muvector}_t) - \R_{S_t}(\muvector)| &<  \Delta_{\min} 
    \tag{Lipschitz  property(Property \ref{prop:lipschitz})}\\
\implies     \R_{S_t}(\overline{\muvector}_t) - \R_{S_t}(\muvector) &< \Delta_{\min} \leq \Delta_{\min}(A_t) \\
    \gamma \cdot \R_{S^{\star}_t}(\overline{\muvector}_t) - \R_{S_t}(\muvector) &< \Delta_{\min}(A_t) \tag{As, $\R_{S_t}(\overline{\muvector}_t) \geq \gamma \cdot \R_{S^{\star}_t}(\overline{\muvector}_t) $} .
\end{align*}
From the  definition of $\Delta_{\min}(A_t)$ and monotonicity property (Property \ref{prop:mono}) we have, a contradiction. 
 Hence, $S_t \notin S_B(A_t)$, which implies that  $\mathbbm{P} \{ S_t \in S_{B}(A_t) \} \leq \frac{2|S_t|}{t^3}$ for all $t \in E$. Hence, $ \mathbbm{P}\{ S_t \in S_{B}(A_t)|B_t \} \leq  2 |S_t|/ t^3.$
\end{proof}
%%%%%%%%%%%%%%%%%%%%%%%%%%%%%%%%%%%%%%%%%%%%%%%%%%%%%%%%%%%%%%%%%%%%%%%%%%%%%%%%%%%%%%%%%%%
\InstIndOne*
\begin{proof}
First, consider the case    $\Delta_{\min} \geq  C \sqrt{\frac{6  k \sigma \log(T)  }{T}}$. From Theorem \ref{thm:Regret_CSUCB_lipschitz} we have,
\begin{align*}
    \mathcal{R}_{\CSUCB}(T) &\leq  \Bigg[  \frac{6C^2 k \sigma \log(T)}{\Delta_{\min}} + 2 k C  \zeta(3) \Big( 1 + \sqrt{3\log(T)/2}\Big)  \Bigg] \\ 
    &\leq C \sqrt{6k\sigma T\log(T)} + 2 k \zeta(3) C (1 + \sqrt{3\log(T)/2}) \tag{as, $\Delta_{\min} \geq C \sqrt{6k\sigma \log(T)/T}$} \\  
    & \leq 3 C \sqrt{6k\sigma T\log(T)} + 2 k C \zeta(3).
\end{align*}
The last inequality follows for all $T \geq k$, from the fact that $\sqrt{6\log(T)} k C \zeta(3) \leq 2\sqrt{6\log(T)}kC \leq 2C \sqrt{6 k \sigma T\log(T) } $. 

Next, let $\Delta_{\min} < C \sqrt{6  k \sigma \log(T) / T}$. Further, let $\eta \geq C \sqrt{6  k \sigma \log(T)/T} $ be a constant. We decompose the regret $\Delta_{S_t}$ at any time $t$ into two parts;  i.e.  $\Delta_{S_t} \geq \eta $ and $\Delta_{S_t}< \eta$, respectively. Thus, instance-independent sleeping regret of \CSUCB, 
\begin{align*}
\mathcal{R}_{\CSUCB}(T) &=  
\mathbbm{E}  \big[  \sum_{t=1}^T \mathbbm{1}( S_t \in S_B(A_t)) \Delta_{S_t} \big ]   = \mathbbm{E}  \big[  \sum_{t=1}^T [\mathbbm{1}( S_t \in S_B(A_t), \Delta_{S_t} < \eta) +   \mathbbm{1}( S_t \in S_B(A_t), \Delta_{S_t} \geq \eta) ] \Delta_{S_t} \big ] . \nonumber
\end{align*}
The first term is upper bounded by $\eta T$. To bound the second term, consider a  \textsc{CSMAB} instance such that $S_{B}^{'}(A) = S_{B}(A) \cap \{S \subseteq A| \Delta_{S_t} \geq \eta \} $. In this instance we have $\Delta_{\max}^{'} = \Delta_{\max}$ and $\Delta_{\min}^{'} = \eta$. Hence, 
\begin{align*}
     \mathcal{R}_{\CSUCB}(T) & \leq \eta T +  \Bigg [ \sum_{t=1}^T \mathbbm{P}\{ S_t  \in  S_{B}^{'}(A_t) \} \Delta_{S_t} \Bigg] \nonumber\\
     & \leq \eta T  + \frac{6C^2 k \sigma \log T}{\Delta_{\min}^{'}}   + 2 k C \zeta(3) ( 1 + \sqrt{3\log(T)/2}) \tag{from Theorem \ref{thm:Regret_CSUCB_lipschitz}}\\
      & \leq \eta T +   \frac{6C^2 k \sigma \log T}{\eta}  + 2 k C \zeta(3) ( 1 + \sqrt{3\log(T)/2}) .
      \tag{As $\Delta_{S_t} \geq \eta$}
\end{align*}
Choose $\eta = C \left(\frac{6 k \sigma \log T }{ T}\right)^{1/2}$ to get the desired upper bound. \end{proof}

%%%%%%%%%%%%%%%%%%%%%%%%%%%%%%%%%%%%%%%%%%%%%%%%%%%%%%%%%%%%%%%%%%%%%%%%%%%%%%%%%%%%%%%%%%%%
\RegLipIndGen*
\iffalse
\begin{restatable}[]{theorem}{RegLipIndGen} 
\label{thm:Regret_CSUCB_independent}
The instance-independent sleeping regret of \CSUCB\ when the reward function satisfies monotonicity (property \ref{prop:mono}) and Lipschitz continuity (property \ref{prop:lipschitz}) properties is given by 
$$ \mathcal\mathcal{R}_{\CSUCB}(T) 
   \leq C(1+ \lambda ) \cdot \sqrt[3]{6k T^2\log (T)} + 2k \lambda C\zeta(3)$$
  where $\lambda = \Big( 1 + \sqrt{\frac{3 \log(T)}{2}}\Big)$.  
\end{restatable}
\fi 
\begin{proof}
%We provide instance independent regret bound for the algorithm where the reward function satisfies monotonicity and Lipschitz continuity properties. We have considered instance independence in terms $\Delta$ terms.
Let the regret of selecting super-arm $S_t$ at round $t$  be, $\Delta_{S_t} := \gamma \cdot \textsc{opt}_{A_t} - \R_{S_t}(\muvector)$, where  $ \textsc{opt}_{A_t}  := \R_{S^{\star}_t}  = \max_{S \subseteq A_t} \R_{S} $. From Theorem \ref{thm:Regret_CSUCB_lipschitz} it is easy to see that the said instance-independent  upper bound holds if $\Delta_{\min} \geq (\frac{T}{\log(T)})^{-1/3}$. Hence, without loss of generality let $\Delta_{\min} < (\frac{T}{\log(T)})^{-1/3}$. Further, let $\eta \geq (\frac{T}{\log(T)})^{-1/3} $ be a constant. We decompose the regret at any time $t$ into two parts, where the per round regret is at most $\eta$ and larger than $\eta$. From Observation \ref{obs:second}, observe that $\Delta_{S_t} \leq C \Big( 1 + \sqrt{3 \log(T)/2}\Big)$. Let $\lambda = \Big( 1 + \sqrt{3 \log(T)/2}\Big)$.
%Without loss of generality, we assume that the reward function $r(.)$ is bounded i.e. there exists $\alpha > 0 $ such that $r_{\boldsymbol{\mu}}(S) \leq \alpha $ for all $\boldsymbol{\mu} \in [0,1]^{k}$ and for all $S \in [k]$.Notice that $\Delta_{\max} \leq \alpha$. 
Thus, instance-independent sleeping regret of \CSUCB, 
\begin{align*}
\mathcal{R}_{\CSUCB}(T) &=   \mathbbm{E}  \big[  \sum_{t=1}^T ( \gamma \cdot \beta \cdot  \R_{S_{t}^{\star}} - \R_{S_t} )) \big ]  \nonumber\\
      & = \Bigg [\sum_{t=1}^T  \mathbbm{P}\{ S_t  \in  S_{B}(A_t)| B_t \} \Delta_{S_t} \mathbbm{1}\{\Delta_{S_t} < \eta \}  + \sum_{t=1}^T \mathbbm{P} \{ S_t  \in  S_{B}(A_t)|B_t \}  \Delta_{S_t}   \mathbbm{1}\{\Delta_{S_t} \geq  \eta \} \Bigg ] \cdot \beta \\
      & \leq \eta T + \sum_{t=1}^T \mathbbm{P} \{ S_t  \in  S_{B}(A_t) \cup  \Delta_{S_t} \geq \eta | B_t \}  \Delta_{S_t} \tag{$\beta \leq 1$}  \nonumber\\
      &=  \eta T + \sum_{t\in T_e \cup E}  \mathbbm{P}\{ S_t  \in  S_{B}(A_t) \cup \Delta_{S_t} \geq  \eta | B_t \} \Delta_{S_t}    + \sum_{t\in D} \mathbbm{P} \{ S_t  \in  S_{B}(A_t) \cup \Delta_{S_t} \geq  \eta | B_t \}  \Delta_{S_t}   \nonumber\\ 
      & \leq  \eta T +  2\zeta(3) k C \Big( 1 + \sqrt{3 \log(T)/2}\Big)   + C \Big( 1 + \sqrt{3 \log(T)/2}\Big) \sum_{t\in D} \mathbbm{P}\{ \Delta_{S_t} \geq \eta \}  \tag{Observation \ref{obs:second}} \nonumber  \\
       & \leq  \eta T +  2k \lambda C\zeta(3)   + C \lambda \sum_{t\in D} \mathbbm{P}\{ \Delta_{S_t} \geq \eta \} \nonumber 
       \\
      & \leq \eta T +  2k \lambda C\zeta(3) + k \lambda C \frac{6C^2\log (T)}{\eta^2}.
\end{align*}
Choose $\eta = C \left(\frac{6 k \log T }{ T}\right)^{1/3}$ to get the following sleeping regret:
$$
      \mathcal{R}_{\CSUCB}(T) 
   \leq C(1+ \lambda ) \cdot \sqrt[3]{6k T^2\log (T)} + 2k \lambda C\zeta(3).
$$
\end{proof}

\begin{comment}
Future work:
\begin{itemize}
    \item Prove lower bound for general reward settings
    \begin{itemize}
        \item for both bounded smoothness and lipschitz setting
        \item Sleeping and non-sleeping case
        \item Fixed size of super-arm and allowing variation in size of super-arms
    \end{itemize}
    
    \item Proof to upper bound for instance independent general reward setting to be of the order $O(\sqrt{T})$. Chen at al. 2013 has showed for a special case of bounded smooth function.
    \begin{itemize}
        \item For sleeping and non-sleeping case
    \end{itemize}
    
    \item Study of How does perturbation in linearity of reward function effects regret?
\end{itemize}
\end{comment}
\RegretBS*
\iffalse
\begin{restatable}[]{theorem}{RegretBS}
\label{thm:Regret_CSUCB1_bounded}
The expected sleeping regret incurred by \CSUCB\ when the reward function satisfies monotonicity (Property \ref{prop:mono}) and bounded smoothness (Property \ref{prop:smooth}) properties is upper bounded by 
$$\mathcal{R}_{\CSUCB} (T) \leq \Big[  \frac{6\log(T)}{(f^{-1}(\Delta_{\min}))^2} +  2\zeta(3) \Big] k \cdot \Delta_{\max}.$$
\end{restatable}
\fi 
\begin{proof}
Following the similar 3 step proof of Theorem \ref{thm:Regret_CSUCB_lipschitz}. We choose with $\ell_t = \frac{6 \log(t)}{(f^{-1}(\Delta_{\min}))^2}$ and $\varepsilon_t = \sqrt{\frac{3 \log(t)}{2 \ell_t}} $ and divide the time instants into sets  $T_e$ and $T_u$ as described in Section \ref{sec:theory results}. Step 1 and 2 is proved as Lemma \ref{lem:BS_Step2} and Lemma \ref{lem:BS_lemmaCase2}. Observe that Step 3 follows trivially as for in Theorem \ref{thm:Regret_CSUCB_lipschitz}. 
\begin{lemma}
\label{lem:BS_Step2}
Let $t \in T_e$, when the reward function satisfies monotonicity and Lipschitz smoothness property then
$
    \mathbbm{P}\{ S_t \in S_{B}(A_t) | B_t\} \leq 2 |A_t| t^{-3}.
$
\end{lemma}
\begin{proof}[Proof of the lemma]   Let $\ell_t := \frac{6 \log(t)}{(f^{-1}(\Delta_{\min}))^2}$ and  $\varepsilon_{t} := \sqrt{\frac{3 \log(t)}{2\ell_t}}$. We have  
\begin{align}
\label{eq:lemma_1}
    \mathbbm{P}\{S_t \in  S_{B}(A_t) \}     
    =  \mathbbm{P}\{ \exists  i \in A_t : | \hat{\mu}_{i,t} - \mu_i |  \geq \varepsilon_t , S_t \in S_{B}(A_t) | B_t \}  
     + \mathbbm{P}\{   \forall i  \in A_t : | \hat{\mu}_{i,t} - \mu_i |  < \varepsilon_t , S_t \in S_{B}(A_t) | B_t \}.
\end{align}
We first prove an upper bound on the first term on the right side of the above expression. We have,  for all the arms $i$ in $A_t$,  
\begin{align*}
     \mathbbm{P}\{|\hat{\mu}_{i,t} - \mu_i|  \geq \varepsilon_t \}  
& \leq 2 e^{- 2 N_{i,t} \varepsilon_t^2} \tag{from Hoeffding's inequality} \\ 
    &  = 2 e^{- N_{i,t} \frac{3 \log(t)}{  \ell_{t}}} \\ 
     & \leq 2 t^{-3}. \tag{as $N_{i,t} \geq \ell_t $}
\end{align*}
Using union bound we get the  following upper bound on the first term  
\begin{align}
\label{eq:lemma1_part1}
   & \mathbbm{P}\big\{ \exists i \in A_t : | \hat{\mu}_{i,t} - \mu_i |  \geq \varepsilon_t , S_t \in \mathcal{S}_{B}(A_t) \big \}  \leq \mathbbm{P} \{ \exists i \in A_t : |\hat{\mu}_{i,t} - \mu_i|  \geq \varepsilon_t \} \leq 2 |A_t| t^{-3}.
\end{align}
Next, we bound the second term. From Equation \ref{eq:8} and the bounded smoothness property  (Property \ref{prop:smooth}), for any $S_t^{'} \subseteq A_t$, we have 
$|\R_{S_t^{'}}(\overline{\muvector}_t) - R_{S_t^{'}}(\muvector) | < f(2 \varepsilon_{t}).$ 
 In particular, for the selected super-arm $S_t$ we have, 
\begin{equation}    |\R_{S_t}(\overline{\muvector}_t) -  \R_{S_t}(\muvector)|  < f(2\varepsilon_t).
\label{eqn:upperBoundExplored}
\end{equation}
This implies,
\begin{align*}
\R_{S_t}(\muvector)  +   \Delta_{\min} & =     \R_{S_t}(\muvector)  +   f(2\varepsilon_t) \tag{As $f(2\varepsilon_t) = \Delta_{\min}$} \\ & > \R_{S_t}(\overline{\muvector}_t) \tag{from Eq.\ref{eqn:upperBoundExplored}}\\
   &\geq \gamma \cdot \R_{S^{\star}_t}(\overline{\muvector}_t) \tag{As $S_t$ is optimal super-arm for $\overline{\muvector}$ } \\ 
   & \geq  \gamma \cdot \R_{S^{\star}_t}(\muvector) = \gamma \cdot \textsc{OPT}_{A_t}.  \tag{from the monotonicity property}
\end{align*}
Hence, we have $\Delta_{\min} > \gamma \cdot \textsc{opt}_{A_t} - \R_{S_t}(\muvector)  $. This contradicts the definition of $\Delta_{\min}$ and hence we have that 
$\mathbbm{P}\big \{ \forall i \in A_t : | \hat{\mu}_{i,t} - \mu_i |  < \varepsilon_t , S_t \in S_{B}(A_t) | B_t \big\} = \mathbbm{P}\big\{ \forall i \in A_t : | \hat{\mu}_{i,t} - \mu_i |  < \varepsilon_t \big\} = 0 $. This, Eq. \ref{eq:lemma_1} and Eq. \ref{eq:lemma1_part1} completes the proof of the lemma.
\end{proof}
\begin{restatable}[]{lemma}{BSLemmaStepThree}
 \label{lem:BS_lemmaCase2}
 For given $t$, if $\forall i \in S_t$, $N_{i,t} \geq \ell_t$ is true and reward function satisfies monotonicity and bounded smoothness property then 
 $$\mathbbm{P}\{S_t \in S_B(A_t)|B_t\} \leq \frac{2|S_t|}{ t^3}  .$$
\end{restatable}
\begin{proof}
Let $\ell_t := \frac{6 \log(t)}{(f^{-1}(\Delta_{\min}))^2}$ and  $\varepsilon_{t} := \sqrt{\frac{3 \log(t)}{2\ell_t}}$.
Consider $t \in E$, where $E= \{t \in T_u | \forall j \in S_t, N_{j,t} \geq \ell_t\}$, i.e., at each $t \in E$ the each arm in super-arm are saturated. 
We have,
\begin{equation}
    \centering
    \R_{S_t}(\overline{\muvector}_t) \geq \R_{S}(\overline{\muvector}_t), \hspace{10pt} \forall S \in 2^{A_t}.
\end{equation}
Let $S^{\star}_t = \argmax_{S \in A_t} \R_{S}(\muvector)$ be an optimal super-arm for given available arms $A_t$ at time $t$. For all $j \in S_t$  we have $N_{j,t}> \ell_t$. Hence from Observation 1, with probability atleast $1 - \frac{2|S_t|}{t^3}$ we have,
\begin{equation}
    \max_{j \in S_t}|\overline{\mu}_{j,t} - \mu_j| \leq 2\varepsilon_{t}.
\end{equation}
%Hence, with probability atleast $\bigg(1 - \frac{1}{t^2}\bigg)^{|S_t|} > 1 - \frac{|S_t|}{t^2}$ we have,
This implies,
\begin{align*}
    |\R_{S_t}(\overline{\muvector}_t) - \R_{S_t}(\mu)| &< \Delta_{\min} 
    \tag{Property \ref{prop:smooth}, $\ell_t$ and $N_{i,t}\geq \ell_t$}\\
\implies     \R_{\overline{\muvector}_t}(S_t) - \R_{S_t}(\muvector) &< \Delta_{\min} \leq \Delta_{\min}(A_t) \\
    \R_{S^{\star}_t}(\overline{\muvector}_t) - \R_{S_t}(\muvector) &< \Delta_{\min}(A_t) \tag{As, $\R_{S_t}(\overline{\muvector}_t) \geq \R_{S_{t}^{\star}}(\overline{\muvector}_t) $} \\
  \implies  \R_{S_t}(\muvector) &> \max_{S \in S_{B}(A_t)} \R_{S}(\muvector). \tag{by definition of $\Delta_{\min}(A_t)$ and monotonicity property}
\end{align*}
Hence, we have $\mathbbm{P}\{S_t \notin S_{B}(A_t)|B_t\} \geq 1 - \frac{2|S_t|}{t^3}$, which implies that $\mathbbm{P}\{S_t \in S_{B}(A_t)|B_t\} \leq  \frac{2|S_t|}{t^3}$ for all $t \in E$. 
\end{proof}

 \textbf{Putting everything together:}

With this, the upper bound on sleeping regret of \CSUCB\ under \BS\ setting is  \begin{align*}
      \mathcal{R}_{\CSUCB}(T) \leq &  \beta \cdot \Delta_{\max} \Big [ \sum_{t\in T_e } \mathbbm{P}\{ S_t \in S_{B}(A_t) | B_t \} +  
     \sum_{t\in E} \mathbbm{P}\{S_t \in S_{B}(A_t)| B_t \}  + \sum_{t\in D} \mathbbm{P}\{S_t \in S_{B}(A_t)| B_t \}  \Big ] \\
     \leq & \beta \cdot \Big [ \sum_{t=1}^T 2\frac{\Delta_{\max}}{ t^3}|S_t| + \Delta_{\max}|D| \Big ] \tag{From Lemma \ref{lem:cardinality of D }, Lemma \ref{lem:BS_Step2}, and Lemma \ref{lem:BS_lemmaCase2}}\\
%      \leq & \beta \cdot \Big [ \sum_{t=1}^T 2 \frac{\Delta_{\max}}{t^3}|S_t| + \Delta_{\max}|D| \Big ]\\
      \leq &  \Big [  2\zeta(3) k \Delta_{\max} + k\ell_T \Delta_{\max} \Big ] \cdot \beta   
      =   \left[  \frac{6C^2\log(T)}{(f^{-1}(\Delta_{\min}))^2} +  2\zeta(3) \right] \beta \cdot k \cdot \Delta_{\max}.
  \end{align*}
\end{proof}

\end{document}